\documentclass[sigconf,nonacm]{acmart}

\renewcommand\footnotetextcopyrightpermission[1]{}
\AtBeginDocument{%
  }

\usepackage{xcolor,colortbl}
\usepackage{booktabs}
\usepackage{pifont}
\usepackage{tabularx}
\usepackage{tikz}
\usepackage{enumitem} 
\usepackage{subcaption}
\usepackage[ruled,vlined]{algorithm2e}
\newcommand*\circled[1]{%
\tikz[baseline=(char.base)]{
  \node[shape=circle, draw=blue, fill=yellow, thick, inner sep=1pt] (char) {\scriptsize\textsf{#1}};
}}
\usepackage{soul}
\usetikzlibrary{shapes.geometric, arrows.meta, positioning, calc}

\begin{document}
\title{Driving Reaction Trajectories via Latent Flow Matching}

\author{Yili Shen}

\affiliation{%
  \institution{University of Notre Dame}
  \city{Notre Dame}
  \state{IN}
  \country{USA}
}

\author{Xiangliang Zhang}
\affiliation{%
  \institution{University of Notre Dame}
  \city{Notre Dame}
  \state{IN}
  \country{USA}
}

\renewcommand{\shortauthors}{}

\begin{abstract}
Recent advances in reaction prediction have achieved near-saturated accuracy on standard benchmarks (e.g., USPTO), yet most state-of-the-art models formulate the task as a one-shot mapping from reactants to products, offering limited insight into the underlying reaction process. Procedural alternatives introduce stepwise generation but often rely on mechanism-specific supervision, discrete symbolic edits, and computationally expensive inference. In this work, we propose \textbf{LatentRxnFlow}, a new reaction prediction paradigm that models reactions as \textbf{continuous latent trajectories} anchored at the thermodynamic product state. Built on \textbf{Conditional Flow Matching}, our approach learns time-dependent latent dynamics directly from standard reactant–product pairs, without requiring mechanistic annotations or curated intermediate labels.
While LatentRxnFlow achieves state-of-the-art performance on USPTO benchmarks, more importantly, the continuous formulation exposes the full generative trajectory, enabling \textbf{trajectory-level diagnostics} that are difficult to realize with discrete or one-shot models. We show that latent trajectory analysis allows us to localize and characterize \textbf{failure modes} and to mitigate certain errors via gated inference. Furthermore, geometric properties of the learned trajectories provide an intrinsic signal of epistemic \textbf{uncertainty}, helping prioritize reliably predictable reaction outcomes and flag ambiguous cases for additional validation. Overall, LatentRxnFlow combines strong predictive accuracy with  transparency, diagnosability, and uncertainty awareness, moving reaction prediction toward more trustworthy  high-throughput discovery workflows.

\end{abstract}



\maketitle

\section{Introduction}
Forward chemical reaction prediction stands as a cornerstone task in the intersection of AI and chemistry, serving as the engine for downstream applications ranging from high-throughput virtual screening to synthetic planning. Recent advances in reaction prediction have led to near-saturated performance on standard benchmarks such as USPTO-50K, USPTO-MIT, and USPTO-Full ~\cite{mclaughlin1983computer,corey1969computer,corey1967general,coley2017prediction,bi2021non,guo2023modeling}, 
where state-of-the-art models routinely achieve over 90\% top-1 accuracy. Most of these approaches formulate reaction prediction as a sequence-to-sequence or graph-to-graph translation problem, operating  on SMILES strings \cite{irwin2022chemformer,schwaller2019molecular,schwaller2018found} or molecular graphs \cite{jin2017predicting,do2019graph,qian2020integrating,sacha2021molecule,bi2021non,tu2022permutation,meng2023doubly,guo2023modeling,keto2025improving,joung2025electron,shen2025towards}.
However, despite their strong empirical performance, most of these methods  operate under a ``one-shot look-ahead'' paradigm: given the reactants, the model directly predicts the final products in a single forward pass.

While effective at learning dataset-level input–output mappings, this formulation diverges from how chemists conceptualize reactivity: \emph{real reactions are kinetic, path-dependent processes driven by continuous physical evolution}, involving electron density redistribution, bond reorganization, and conformational changes over time. 
Collapsing this dynamics into a single-step prediction can therefore limit mechanistic transparency: the model may output the final products but provides little information about plausible intermediates, or competing pathways. As a result, it is often difficult to diagnose \emph{why} a prediction fails, to understand \emph{how} different reaction mechanisms or pathway choices influence the outcome, or to assess whether the model's \emph{confidence} reflects genuine chemical certainty versus dataset-level correlations.


To explicitly model the intermediate transformation, recent procedural models such as GTPN~\cite{do2019graph}, MEGAN~\cite{sacha2021molecule}, and FlowER~\cite{joung2025electron} decompose  the reaction into a sequence of elementary steps or graph-edit operations. 
This formulation introduces a temporal dimension by ordering bond-breaking and bond-forming events into a stepwise progression, defined either by rule-derived graph edits (as in GTPN and MEGAN) or by mechanistic elementary step (as in FlowER), allowing the model to represent how molecular structures evolve from reactants toward products rather than predicting the outcome in a single jump. At the same time, it provides finer granularity by operating at the level of localized structural changes (e.g., individual bond edits or atom-level transformations), instead of treating the reaction as an atomic input–output mapping.
However, translating these procedural advantages into large-scale applications remains challenging due to multiple practical bottlenecks.

\textbf{First, reliance on high-fidelity mechanistic labels.} To enable a stepwise reaction procedure, methods such as FlowER~\cite{joung2025electron} leverage detailed mechanistic supervision (e.g., mechanism templates or pathway constraints). However, such information is typically absent from widely used reaction datasets (e.g.,   USPTO). It thus requires utilizing commercial classification tools (e.g., NameRxn) for template extraction or expert input to define mechanistic rules and pathway constraints. 
Furthermore, effective training commonly assumes strictly curated reaction records with explicit hydrogens, precise atom-mapping, and complete byproduct annotations, which are difficult to obtain reliably from standard, noisy public benchmarks (e.g., USPTO).
Consequently, this reliance on curated, mechanism-rich data can reduce the feasibility and flexibility of applying such models in large-scale or real-world settings where such annotations are unavailable or inconsistent.

\textbf{Second, limited mechanistic grounding under weak supervision.}
In contrast to mechanism-supervised methods, approaches such as GTPN~\cite{do2019graph} and MEGAN~\cite{sacha2021molecule} infer reaction pathways under weaker supervision, representing reactions as sequences of graph edits learned primarily from end-point reactant--product pairs. While this avoids the need for explicit mechanistic labels, the resulting edit trajectories may not correspond to physically grounded reaction coordinates. Consequently, the inferred ``pseudo-paths'' may deviate from chemically plausible mechanisms in some cases, limiting their reliability for rigorous mechanistic interpretation and pathway-level analysis.

\textbf{Third, computational overhead in autoregressive decoding}. 
In autoregressive  approaches (e.g., {SMILES-based approaches ~\cite{nam2016linking,schwaller2018found,irwin2022chemformer,lu2022unified, tu2022permutation}}), a reaction is generated as a long chain of intermediate steps, often exceeding 10 actions. Identifying a high-probability trajectory typically requires explicit search (e.g., beam search) over a combinatorial space of candidate sequences ({token-level in SMILES-based approaches and {intermediate}-level in step-wise procedural generation approaches}), which can substantially increase runtime. Repeatedly expanding partial trajectories and applying step-level constraints or validity checks increases both memory consumption and inference latency compared with direct end-point prediction.

Motivated by these limitations, we propose to build a new reaction prediction paradigm that is \textbf{data-efficient} (learning only from standard Reactant-Product pairs), \textbf{dynamics-transparent} (enabling verification by exposing interpretable intermediate evolution), and \textbf{uncertainty-informative} (enabling post-hoc confidence estimation from inference trajectories). 
To this end, we adopt  \textbf{Conditional Flow Matching (CFM)}~\cite{dao2023flow,lipman2022flow,tong2023conditional,liu2022flow}, an emerging framework for learning continuous-time generative dynamics that has shown promise in scientific domains (e.g., modeling molecular transformations and transition-state–related structures)~\cite{duan2025optimal,luo2025generating,zeng2026propmolflow,dunn2024mixed,zhou2025energy}. CFM provides a principled way to move beyond a purely static reactant$\rightarrow$product mapping by learning a \textbf{time-dependent latent vector field} that transports samples from a reactant distribution to a product distribution. This \textbf{latent continuous} formulation better reflects the \textbf{kinetic, path-dependent nature of reactions} than one-shot predictors, which directly regress the final outcome without an explicit notion of intermediate evolution.

\begin{figure}[t]
    \centering 
    \includegraphics[width=0.6\textwidth]{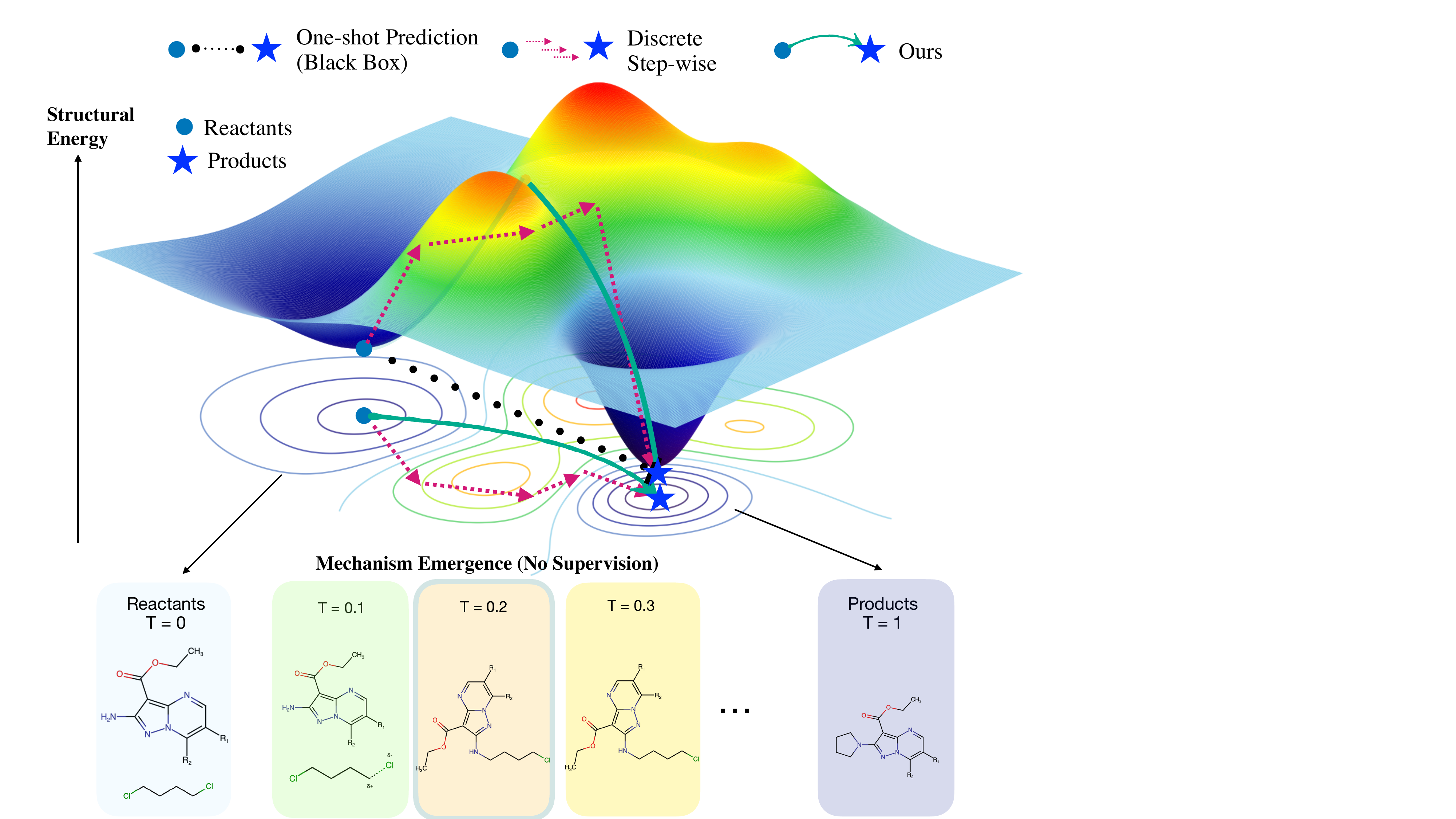} 
    \caption{\textbf{Paradigm Shift: Continuous Latent Dynamics for Reaction Prediction.} 
    Unlike discrete graph edits or one-shot black-box prediction, \textit{LatentRxnFlow} models chemical transformations as continuous trajectories on a learned structural manifold. 
    The landscape visualizes the effective potential energy in the latent space. 
    Notably, this continuous dynamics approach allows for the spontaneous emergence of reaction mechanisms, where the trajectory passes through chemically valid intermediate states (bottom panels), offering a lens beyond simple input-output mapping.} \vspace{-0.25in}
    \label{fig:teaser}
\end{figure}

Concretely, we learn a vector field $v_{\theta}(z|\cdot)$ in a \textbf{latent space} and simulate a reaction as a \textbf{smooth transport process}: the reactant representation $z_r$ evolves continuously along a reaction coordinate toward a product state $z_p$. This enables intermediate states $z_t$ that can be inspected or checked for consistency, and avoids committing to a discrete sequence of symbolic edits at every step. As a result, the model can reduce stepwise error accumulation common in autoregressive edit-based procedures, while still producing discrete products via decoding only at the end of the continuous trajectory.
In contrast, FlowER~\cite{joung2025electron} applies flow-matching objectives directly to discrete bond matrices, so generation proceeds in a discrete, edit-like state space. Training such a generator typically relies on mechanism-derived constraints and carefully curated structural annotations to ensure valid transitions. As a result, uncertainty and alternative pathways are  collapsed into a single discrete edit at each step, because the model must commit to a specific edit before progressing further, as illustrated in Fig. \ref{fig:teaser}.

This difference between discrete edit-based generation and continuous latent-trajectory modeling can be illustrated by a river-crossing analogy. Discrete procedural models resemble crossing a river by jumping across stepping stones: at each step, the model must select and commit to a single stone (edit) before proceeding, thereby discarding other plausible options. These choices are largely irreversible, so an early misstep can propagate and constrain subsequent decisions. In contrast, learning a latent continuous trajectory is akin to moving through a shallow, continuous current: the path evolves smoothly, nearby routes remain accessible, and gradual course corrections are possible. This continuous evolution preserves information about alternative pathways and uncertainty throughout the process, rather than collapsing it into a sequence of discrete commitments. 

Then, a natural question is whether these two paradigms can achieve comparable predictive accuracy. We evaluate our proposed model, named LatentRxnFlow (which models reactions as continuous latent trajectories anchored at the thermodynamic product state) on standard USPTO benchmarks. LatentRxnFlow achieves state-of-the-art performance (e.g., over 90\% Top-1 accuracy), matching or exceeding strong one-shot baselines while learning directly from standard reactant–product pairs. Beyond endpoint accuracy, our continuous formulation enables capabilities that are difficult to realize with one-shot or discrete procedural baselines:
\begin{itemize}[leftmargin=*]
    \item \textbf{Diagnosable generation and failure rectification.} Unlike one-shot models, which fail silently by emitting an incorrect product with little indication of how the error arose, \textbf{LatentRxnFlow exposes a continuous generative trajectory that can be inspected as it evolves from reactants toward products}. Concretely, we can project latent states along the trajectory into 2D/3D (e.g., via PCA or UMAP) to visualize the evolution, and localize when and how the trajectory begins to deviate from the desired  path. In particular, we can pinpoint the onset of failure modes such as \emph{overshooting}, where the trajectory passes through (or very near) the correct product region yet continued evolution drives it into an incorrect basin.  {Example   observations are demonstrated in Fig. \ref{fig:comparison}  and  Fig. \ref{fig:traj_viz}.}  This observability enables \textbf{actionable intervention}: errors can be mitigated by halting or gating the flow before divergence, preventing certain failures rather than merely detecting them post hoc.
    
    \item \textbf{Latent geometry as an uncertainty signal.} Flow Matching encourages trajectories that approximate straight-line optimal transport paths between reactant and product distributions. Empirically, we observe that confident predictions correspond to smooth, low-curvature trajectories, whereas ambiguous or complex reactions induce more curved or unstable paths (see Fig. \ref{fig:traj_viz}).  These \emph{trajectory-geometry cues}  provide a principled proxy for epistemic uncertainty, helping prioritize reliably predictable reaction outcomes and flag ambiguous cases for additional validation. Notably, this signal is derived directly from the learned dynamics, without requiring   repeated sampling. 
\end{itemize}

Overall, LatentRxnFlow combines strong predictive accuracy with trajectory-level transparency, actionable diagnosability, and intrinsic uncertainty signals, advancing reaction prediction toward more trustworthy  high-throughput discovery workflows.

\begin{table*}[t]
\centering
\caption{Comparison of reaction prediction paradigms and their capabilities beyond outcome prediction.} \vspace{-0.1in}
\label{tab:paradigm_comparison}
\resizebox{0.89\textwidth}{!}{%
\begin{tabular}{l|ccc}
\toprule
\textbf{Models} & \textbf{1-shot Prediction} & \textbf{Procedural Prediction} & \textbf{Latent Continuous Prediction} \\
 & \small{(Chemformer/NERF/etc)} & \small{(FlowER/MEGAN)} & \small{(Our LatentRxnFlow)} \\
\midrule
\textbf{Prediction Paradigm} & Direct endpoint mapping & Procedural action sequence &  Dynamic continuous process \\
\textbf{Prediction Space} & Product tokens/bonds configuration & Discrete graph edit actions & {Continuous latent trajectory} \\
\textbf{Intermediate States} & \ding{55} (Undefined) & Edited graphs (rule-constrained) & {Latent state (decodable)} \\
\textbf{Inference Mechanism} & Autoregressive decoding (beam search) & Step-wise rollout & Continuous flow integration \\
\midrule
\textbf{Failure Analysis Support} & None (endpoint only) & Limited (step-wise errors) & Explicit (trajectory-level diagnosis) \\
\textbf{Online Error Mitigation}   & None  & Manual/heuristic rules & Adaptive gating \\
\textbf{Ambiguity flagging} & None & None & Geometry-based uncertainty signal \\
\bottomrule
\end{tabular}%
}
\end{table*}


\section{Related Work}
\label{sec:related}
\subsection{Reaction Prediction}
We categorize existing approaches to reaction prediction by how  representing molecular structure and whether  modeling reactions as static endpoint mappings or explicit transformation processes. 
\vspace{-0.08in}
\subsubsection{SMILES-based Sequence Modeling.} A major class of existing work treats reaction prediction as a machine translation task, taking the SMILES of reactants as input and generating the SMILES of products as output. Early methods relied on sequence-to-sequence (seq2seq) architectures with RNN-based encoders and decoders~\cite{nam2016linking,schwaller2018found}, while subsequent work shifted to Transformer-based autoregressive models ~\cite{schwaller2019molecular,tetko2020state,irwin2022chemformer,lu2022unified}.
These methods have demonstrated strong performance on large-scale benchmarks such as USPTO~\cite{jin2017predicting,schneider2016s}. However, these models operate as "black boxes", without explicitly modeling the chemical topology or the reaction process, limiting their interpretability.

\vspace{-0.08in}
\subsubsection{Graph-based One-Shot Prediction.} 
By explicitly modeling atom connectivity and molecular topology, graph-based frameworks such as NERF~\cite{bi2021non} and ReactionSink~\cite{meng2023doubly} formulate reactions as electron redistributions, and predict bond changes in a one-shot manner (e.g., using multi-pointer networks or identifying electron sinks). NERF~\cite{bi2021non} pioneered this by providing interpretable signals related to electron flow, and ReactionSink~\cite{meng2023doubly} further incorporates physical constraints to improve plausibility. While these models offer substantial inference speedups (e.g., NERF reports a 27$\times$ speedup over autoregressive Transformers~\cite{bi2021non}), they fundamentally remain thermodynamic endpoint predictors. They predict the final electron configuration directly from the input, skipping the dynamic evolution of the reaction, thus lacking the ability to diagnose how the transformation unfolds over time.

\vspace{-0.1in}
\subsubsection{Procedural and Step-wise Generation.} To introduce 
interpretability beyond endpoint prediction,
procedural models like MEGAN~\cite{sacha2021molecule}, GTPN~\cite{do2019graph}, and the recent FlowER~\cite{joung2025electron} formulate reaction prediction as a sequence of graph edits or elementary steps. For instance, MEGAN utilizes reinforcement learning to predict a sequence of graph actions (Add/Delete Node/Bond). However, these discrete procedural approaches face critical bottlenecks: (1) Error Propagation: 
a single failed step in the successive process can invalidate the entire synthesis outcome. (2) Scalability: models like FlowER rely on expensive, expert-curated mechanism data, MEGAN and GTPN require beam search over a combinatorial action space, making them computationally prohibitive for large-scale screening compared to one-shot models.

\textbf{LatentRxnFlow} matches the efficiency of one-shot predictors by learning only from standard reactant--product pairs, while providing process interpretability through an explicit transformation trajectory, and it avoids the step-wise error propagation that often affects discrete edit-based procedural models. We provide a comparison of \textbf{LatentRxnFlow} with existing approaches in Table~\ref{tab:paradigm_comparison}, highlighting key differences in modeling paradigm, and the resulting capabilities beyond outcome prediction.

\vspace{-0.15in}
\subsection{Continuous Generative Models}
Generative modeling for scientific data has evolved rapidly in recent years. Focusing on diffusion- and flow-based methods, this line of work has progressed from early successes with Denoising Diffusion Probabilistic Models (DDPMs)~\cite{ho2020denoising} and score-based generative models~\cite{song2020score} to deterministic continuous-time formulations such as Continuous Normalizing Flows (CNFs)~\cite{chen2018neural}. Flow Matching (FM)~\cite{lipman2022flow,tong2023conditional,liu2022flow} recently emerged as a practical breakthrough, enabling simulation-free training of CNFs by directly regressing a target vector field. By constructing deterministic Optimal Transport (OT) paths between distributions, FM yields fast and stable generation, and is particularly appealing for scientific settings where smooth, analyzable dynamics are desirable.

Recent work   FlowER~\cite{joung2025electron} has explored applying FM to reaction prediction, but does so in a fundamentally different manner. FlowER operates directly in \emph{discrete bond space}, constructing supervision by linearly interpolating between reactant and product adjacency matrices. This induces intermediate states that correspond to fractional or partially formed bonds, which lack clear physical meaning, and enforces a uniform rate of structural change that does not align with the non-linear, barrier-driven nature of chemical reactions.

In contrast, our approach applies Flow Matching in a \emph{continuous latent manifold}, where interpolation occurs between compact latent representations rather than discrete molecular graphs. Although the latent trajectory follows a linear OT path, the non-linear decoding back to molecular structure induces non-linear observable dynamics, allowing the model to capture sharp structural transitions after extended periods of stability. This design aligns more naturally with reaction kinetics, enabling the model to represent gradual energy accumulation followed by rapid transformation—without requiring explicit transition-state annotations or mechanistic supervision.

\section{Methodologies}
\label{sec:method}

\textbf{Problem Formulation: Reaction as Topological Redistribution.}
Given a reactant molecular graph $G_r = (\mathcal{V}, \mathcal{E}_r)$ with atoms $\mathcal{V}$ and bonds $\mathcal{E}_r$, the goal of reaction prediction is to predict the product graph $G_p = (\mathcal{V}, \mathcal{E}_p)$. Since standard chemical reactions preserve the atom mapping (mass conservation), the atom set $\mathcal{V}$ remains invariant. Therefore, the task effectively reduces to predicting the topological redistribution of bonds and the state update of atom properties. Formally, we define the product prediction as:
\begin{equation} \label{eq:problem-setting}
\mathcal{E}_p = \mathcal{E}_r + \Delta \mathcal{E}, \quad Q_p = f_Q(z), \quad A_p = f_A(z)
\end{equation}
where $\Delta \mathcal{E}$ represents the explicit bond changes (breakage and formation), while $Q_p$ (formal charge) and $A_p$ (aromaticity) are predicted directly from the learned representation $z$ of reactant. This hybrid formulation combines the data efficiency of residual learning (for sparse bond changes) with the flexibility of direct prediction (for atom properties).

\vspace{-0.05in}
\subsection{The LatentRxnFlow Framework}
\begin{figure}[t]
  \centering
  \includegraphics[width=1.05\linewidth]{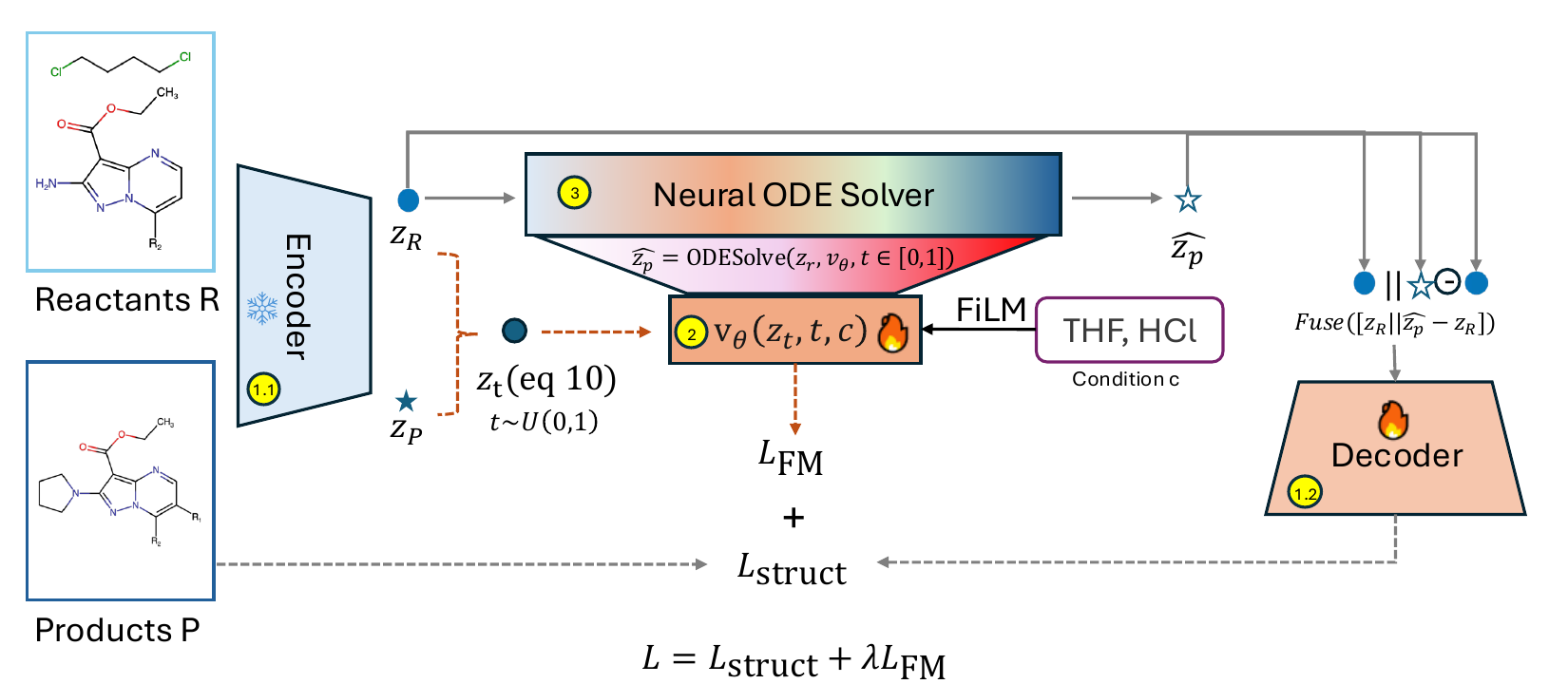} 
  
  \vspace{-0.2cm} 
  
  \caption{\textbf{The LatentRxnFlow Framework. Solid lines denote operations executed   during both inference time and training time; Dash lines  indicate training-only operations. } 
  }
  \label{fig:framework}
  \vspace{-0.3cm} 
\end{figure}

Our proposed framework consists of three coupled components: \circled{1} A Graph Autoencoder (\textbf{GAE}) backbone that compresses discrete structures into continuous embeddings and reconstructs structural updates ($\Delta \mathcal{E},Q_p,A_p$); \circled{2} A Conditional Flow Matching (\textbf{CFM}) module that learns the reaction kinetics (vector field $v_\theta$) in the latent space; and \circled{3} An  Ordinary Differential Equation (\textbf{ODE}) based inference engine for producing $\hat{z}_p$, as shown in Fig. \ref{fig:framework}.


\vspace{+0.03in}
\noindent
\circled{1.1} \textbf{Encoder of GAE.} 
This transformer-based encoder with edge-aware message passing   maps the reactant $G_r$  to a continuous node embedding matrix $z_r \in \mathbb{R}^{|\mathcal{V}| \times d}$.

\vspace{+0.03in}
\noindent
\circled{1.2} \textbf{Decoder of GAE.} 
Taking an embedding  $z$ (which could be $z_r$,  the evolved $z_t$ or the predicted $\hat{z}_p$),  the decoder reconstructs the product structure using a hybrid strategy:
\begin{itemize}[leftmargin=*]
\item \textbf{Residual Bond Decoding ($\Delta \mathcal{E}$):} 
Following Eq. (\ref{eq:problem-setting}), we
predict a differential update to the reactant graph using dual attention for bond formation (\textit{increments} $M_\text{inc}$), and bond breakage (\textit{decrements}  $M_{dec}$), both 
  from   $z$. 
The predicted product bond matrix $\hat{A}_p$ is:
\begin{equation} \label{eq:residual-bond}
\hat{A}_p = A_r + (M_\text{inc} - M_{dec})
\end{equation}
This differential design acts as a strong inductive bias, ensuring the model defaults to the reactant structure ($A_r$) and focuses its capacity solely on the active reaction center.
\item \textbf{Direct Property Prediction ($Q_p, A_p$):} For atom-level properties such as \emph{Formal Charge} and \emph{Aromaticity}, we employ linear heads directly on the latent features  $z$ to classify the final state of each atom (e.g., charge $\in \{-6, \dots, +6\}$).
\end{itemize}

\vspace{+0.03in}
\noindent
\circled{2} \textbf{CFM in Latent Space.}
To model the reaction kinetics, we insert a CFM module between the encoder and decoder  and learn a time-dependent vector field
$v_\theta: \mathcal{Z} \times [0, 1] \times \mathcal{C} \to \mathbb{R}^d$ that drives the reactant embedding $z_r$ (source, $t=0$) to the product embedding $z_p$ (target, $t=1$) conditioned on reaction context $c$.
The latent state $z_t$ is advanced by  integrating the ODE:
\begin{equation}\label{eq:ode}
\frac{d z_t}{d t} = v_\theta(z_t,t,c),
\end{equation}
producing a \textbf{continuous trajectory} $\{z_t\}_{t\in[0,1]}$ that deterministically transports   toward the product embedding.

\vspace{+0.03in}
\noindent
\circled{3} \textbf{ODE Integration for Inference (Reaction Prediction).}
After training, given a reactant embedding $z_r$ and condition vector $h_c$, the product is inferred by a fixed-step ODE solver (e.g., RK4 or Heun) with $N$ steps to approximate the solution:
\begin{equation}
\hat{z}_p \approx z_r + \sum_{i=0}^{N-1} v_\theta(z_{t_i}, t_i, h_c)\,\Delta t,
\qquad t_i=\frac{i}{N},\ \Delta t=\frac{1}{N}.
\end{equation}
The resulting $\hat{z}_p$, along with $z_r$, goes through Scaffold-Anchored Residual Fusion (eq. ~\ref{eq:fuse}), is then decoded by \circled{1.2} Decoder of GAE. 

We next describe how the reaction context $c$ is incorporated into the model and how the time-dependent vector field $v_\theta$ is implemented as a neural network.

\subsection{Reaction Condition Encoding}
In chemical reactions, auxiliary components such as solvents, catalysts, and reagents do not contribute atoms to the final products, yet they shape the reaction environment and  influence the reaction pathway. We refer to them collectively as \emph{conditional agents} $\mathcal{C}$. Encoding $\mathcal{C}$ should consider the \emph{long-tail} distribution of chemical reagents: a small number of agents appear frequently, while most are rare or specialized. We encoder them by different strategies. 

\noindent
\textbf{Frequent-condition indicator ($h_{\text{freq}}$).}
  We identify the top-$K$ most frequent conditional agents in the training corpus (e.g., common solvents such as THF and DCM). For a given reaction, we construct a $K$-dimensional multi-hot vector indicating the presence of these high-frequency agents. This captures strong, explicit priors associated with common reaction environments.

\noindent 
\textbf{Condition-set semantic embedding ($h_{\text{set}}$).}
To represent the full diversity of conditional agents (including rare catalysts), we compute molecular fingerprints (e.g., 512-bit ECFP4) for each molecule $m \in \mathcal{C}$. Since the number and order of agents vary across reactions, we treat $\mathcal{C}$ as a permutation-invariant set and apply a DeepSets-style encoder:
\begin{equation}
    h_{\text{set}} \;=\; \text{Pool}\Bigl(\bigl\{\, \text{MLP}_{\text{fp}}\bigl(\text{FP}(m)\bigr) \;\big|\; m \in \mathcal{C} \,\bigr\}\Bigr),
\end{equation}
where $\text{Pool}(\cdot)$ is a permutation-invariant aggregation operator (e.g., gated attention pooling). This component provides a semantic representation that generalizes across diverse and infrequent reagents.

The final condition vector is the concatenation
\begin{equation}
  h_c \;=\; \bigl[\, h_{\text{freq}},\; h_{\text{set}} \,\bigr],
\end{equation}
which integrates explicit frequency-based priors with chemically meaningful semantic information.

\subsection{Vector-Field $v_\theta$ Parameterization} 

\vspace{+0.03in}
\noindent
\textbf{Baseline field}.
The core of our generative engine is a neural vector field $v_\theta(z_t, t, h_c)$.
To capture both intrinsic scaffold reactivity and extrinsic conditional effects, we first learn an \textbf{\emph{unconditional}} baseline field that models the reactant-driven transformation, and then \textbf{\emph{inject reaction conditions}}.
Concretely,  we encode the continuous time variable $t \in [0,1]$ via sinusoidal positional embeddings to produce $e_t$, enabling the model to distinguish different phases of the transformation (e.g., the early "activation"  and late "relaxation" phase of the reaction). A backbone network $\Phi_{\text{base}}$ processes the concatenated input $[z_t, e_t]$ to produce a baseline velocity field:
\begin{equation}
v_{\text{base}} = \Phi_{\text{base}}([z_t, e_t]).
\end{equation}


\vspace{+0.03in}
\noindent
\textbf{Conditional Injection:  Residual Feature-wise Linear Modulation (FiLM).}
To incorporate the condition vector $h_c$ into $v_{base}$, we use residual FiLM, a lightweight global modulation (avoiding heavy cross-attention for efficiency).
We hypothesize that reaction conditions (e.g., catalysts, solvents, temperature) act as \emph{global fields} that reshape the effective Potential Energy Surface (PES), modulating \emph{rates and selectivity}, without altering the fundamental atom mapping. Consistent with this view, the condition vector $h_c$ is mapped to a scale vector $\gamma$ and a shift vector $\beta$:
\begin{equation}
\gamma, \beta = \text{MLP}_{\text{cond}}(h_c).
\end{equation}
The  vector field is then defined as
\begin{equation}
v_\theta(z_t, t, h_c) = v_{\text{base}} \odot (1 + \gamma) + \beta,
\end{equation}
where $\odot$ denotes element-wise multiplication. Intuitively, the scaling factor $\gamma$ modulates the magnitude of structural change (condition-dependent \emph{rate effects} like a catalyst lowering the activation energy, effectively "speeding up" the flow), 
while the shifting factor $\beta$ introduces directional bias (condition-dependent \emph{selectivity} like a specific reagent favoring one stereoisomer over another).

\subsection{End-to-End Training}
We train the entire \textbf{LatentRxnFlow} framework end-to-end {to jointly optimize the vector field $v_\theta$, the decoder of GAE, and the  encoder of condition}. The overall objective combines a \textbf{flow matching loss} for learning \textbf{latent reaction dynamics} with a   \textbf{reconstruction loss} for structural \textbf{decoding}. This design ensures that the inferred latent trajectories   are both kinetically meaningful and decodable into valid molecular graphs.

\vspace{+0.04in}
\noindent
\textbf{Flow Matching Loss.}
Given reactant--product pairs $\{z_r, z_p\}$ from the training data, we supervise the vector field $v_\theta$ using intermediate latent states sampled from a stochastic probability path centered on the Optimal Transport (OT) trajectory:
\begin{equation}
z_t = (1 - t) z_r + t z_p + \sigma \epsilon, 
\qquad \epsilon \sim \mathcal{N}(0, I),
\end{equation}
where $\sigma$ controls the width of the training manifold. This ``thickened'' path exposes the vector field to a local neighborhood around the ideal OT path, improving robustness to numerical integration errors and modeling imperfections. 
 {Notably, the ODE solver \circled{3} employed at inference time  is \emph{not} used in constructing Flow Mathing Loss,} because  the ground-truth $z_p$ is available during training; we can thus analytically sample     $z_t$ along a simple path between $z_r$ and $z_p$ and directly supervise the vector field, without numerically integrating the dynamics.  
  
The flow matching objective minimizes the discrepancy  over sampled time steps $t \sim \mathcal{U}(0, 1)$ 
\begin{equation}
\mathcal{L}_{\text{flow}}
=
\mathbb{E}_{t, z_r, z_p, \epsilon}
\left[
\left\| v_\theta(z_t, t, h_c) - (z_p - z_r) \right\|_2^2
\right].
\end{equation}
Minimizing this loss encourages $v_\theta$ to learn the latent driving force that transports $z_r$  toward $z_p$ under   condition $h_c$.
 
\vspace{+0.04in}
\noindent
\textbf{Structural Reconstruction Loss.}
The GAE decoder maps the flow-generated latent embedding to the product structure by predicting  topological bond redistributions and atom-level properties, following Eq.~(\ref{eq:problem-setting}).   
 We therefore minimize the reconstruction loss:
\begin{equation}
\mathcal{L}_{\text{struct}}
=
\mathcal{L}_{\text{bond}}\!\left(\Delta \mathcal{E}_{\text{pred}}, \Delta \mathcal{E}_{\text{gt}}\right)
\;+\;
\lambda_{\text{prop}}\left(\mathcal{L}_{\text{charge}} + \mathcal{L}_{\text{aroma}}\right),
\end{equation}
where $\mathcal{L}_{\text{bond}}$ measures the discrepancy between  the ground truth  with the predicted bond changes   $\Delta \mathcal{E}_{\text{pred}} = M_{\text{inc}} - M_{\text{dec}}$  (in Eq.~\ref{eq:residual-bond}). 
$\mathcal{L}_{\text{charge}}$ and $\mathcal{L}_{\text{aroma}}$ are cross-entropy losses for atom-level properties (formal charge and aromaticity classified in Module~\circled{1.2}), and $\lambda_{\text{prop}}$ balances bond and property supervision.



\noindent
\emph{Decoder input.} 
To align the decoder's input distribution with inference-time dynamics, we adopt a \textbf{solver-in-the-loop} training strategy that explicitly simulates inference during training (using ODE to produce the decoder input $\hat{z}_p$). Concretely, for each batch, we integrate the learned vector field starting from the reactant embedding $z_r$ to obtain
\begin{equation}
\hat{z}_p = \text{ODESolve}(z_r, v_\theta; s \in [0,1]).
\end{equation}
Here, $\text{ODESolve}$ can be a differentiable solver (e.g., Heun with $N=5$ steps).
Since numerical integration can accumulate error and cause $\hat{z}_p$ to drift, we introduce \textbf{Scaffold-Anchored Residual Fusion}, using $z_r$ as a stable structural anchor. Specifically, rather than decoding from $\hat{z}_p$ directly, we fuse the anchor with the predicted displacement:
\begin{equation}\label{eq:fuse}
z_{\text{in}} = \text{MLP}_{\text{fuse}}\!\left([\, z_r \mathbin{\|} (\hat{z}_p - z_r) \,]\right),
\end{equation}
where $\mathbin{\|}$ denotes feature concatenation. This representation provides the decoder with (i) the intact scaffold $z_r$ and (ii) a residual update $(\hat{z}_p - z_r)$ that isolates reaction-induced changes, encouraging sparse bond edits rather than reconstructing the molecule from scratch. In practice, this improves structural validity and robustness to solver-induced deviations.

The final loss function is a weighted sum:
\begin{equation}\mathcal{L}_\text{total} = \mathcal{L}_\text{struct} + \lambda_\text{flow}. \mathcal{L}_\text{flow}.
\end{equation}

\section{Experiments}
\label{sec:experiments}
\textbf{Dataset and Baselines.} We evaluate LatentRxnFlow on the standard USPTO-480k benchmark, consisting of $\sim$480k reaction data extracted from patent dataset. We adopt the canonical 400k/40k/40k split for training, validation, and testing to align with the benchmark settings of previous work. We compare our approach against strong baselines across three categories:  (1) Sequence-based
: Molecular Transformer~\cite{schwaller2019molecular}, Chemformer~\cite{irwin2022chemformer}, Graph2SMILES~\cite{tu2022permutation};  (2) Graph-based explicit discrete trajectories: MEGAN~\cite{sacha2021molecule}, GTPN~\cite{do2019graph}, FlowER~\cite{joung2025electron}, and (3) Non-autoregressive: WLDN~\cite{jin2017predicting}, NERF~\cite{bi2021non}. We use the standard Top-k accuracy as the primary metric. Please refer to Appendix ~\ref{app:impl_details} for more implementation details.



\subsection{Main Results: Comparative Performance}
Table \ref{tab:main_results} summarizes the overall performance:
LatentRxnFlow achieves a Top-1 accuracy of 90.2\%, making it competitive with the strongest  baseline NERF.
We also evaluate two  variants to isolate the contributions of key components.  (i) \textbf{w/o Flow (Direct Decode)} removes the latent flow module between the encoder and decoder, and directly decodes from the encoder representation;
(ii) \textbf{w/o FiLM (No Conditioning Injection)} disables FiLM-based conditional injection in CFM. {We concatenate the condition molecules in the reactant input end instead.}
Both variants incur accuracy degradation, most pronounced for Top-$k$ ($k$>1), implying the usefulness of these designs in the framework. 
Due to space limit, more  performance analysis including ablation studies can be found in Appendix ~\ref{app:overall-res}.

\subsection{Efficiency Analysis}\label{sec:efficiency}

We analyze the computational efficiency of LatentRxnFlow from two perspectives: (1) a theoretical complexity derivation comparing conditioning strategies (Appendix \ref{app:Efficiency}), and (2) an empirical comparison against discrete generative baselines:  we benchmark latency using the robust RK4 solver. As shown in Table \ref{tab:efficiency_comparison}, LatentRxnFlow achieves the lowest latency (8.5 ms). Notably, increasing the RK4 integration steps from 10 to 100 incurs negligible cost ($\sim$0.1ms), confirming that inference is dominated by other overhead rather than the lightweight ODE solving. See more in Appendix \ref{app:Efficiency}.

\subsection{Failure Diagnosis} 
\label{sec:trajectory_diagnosis}

\begin{figure}[t]
  \centering
  \begin{subfigure}[b]{0.58\linewidth}\label{fig:demo} 
    \includegraphics[width=\linewidth]{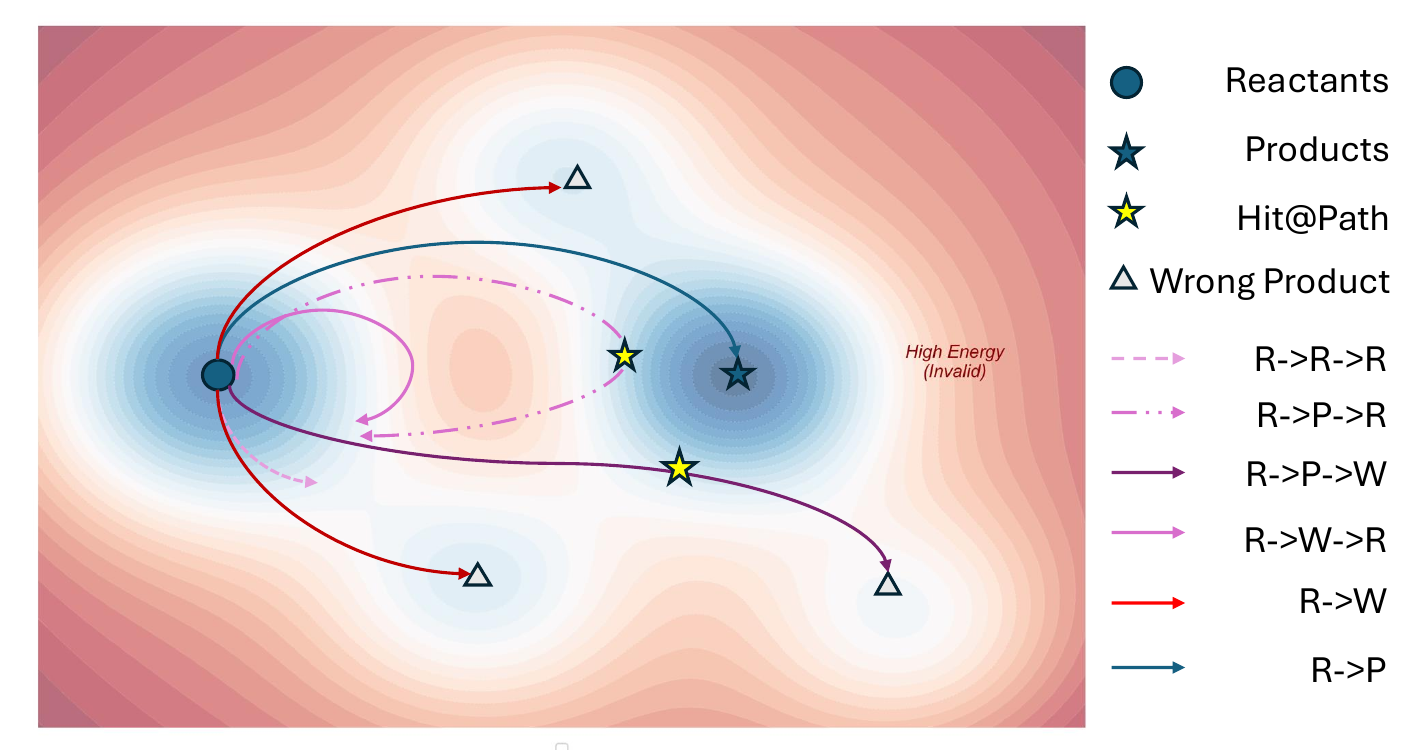} 
    
  \end{subfigure}
  \hfill 
  \begin{subfigure}[b]{0.4\linewidth}
    \includegraphics[width=\linewidth]{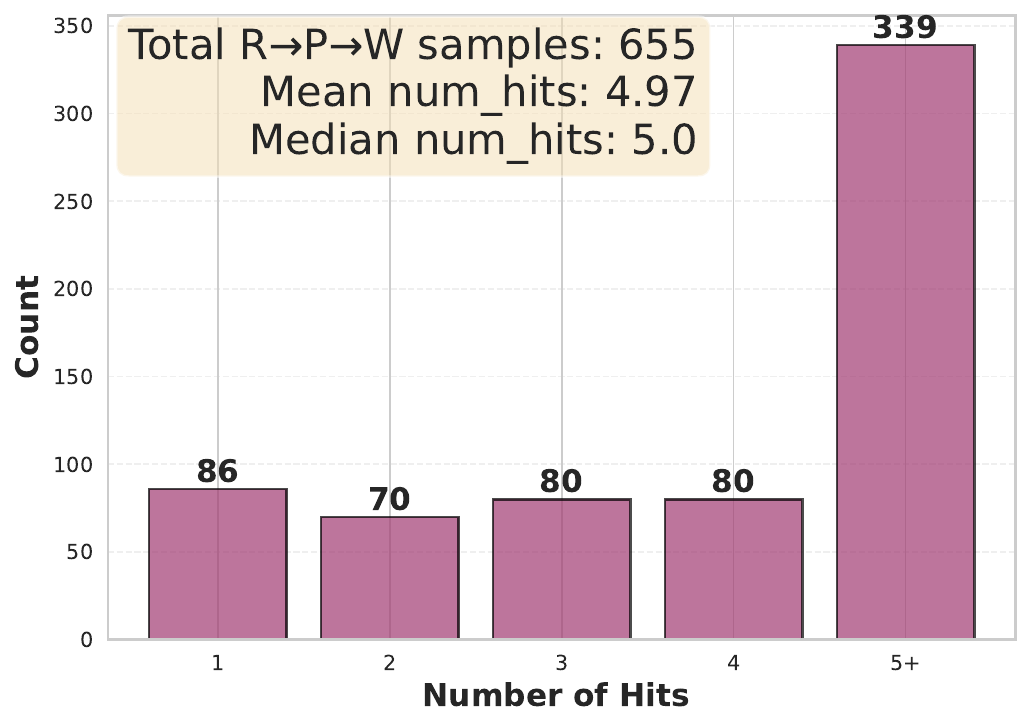} 
    \label{fig:hit_num}
  \end{subfigure}
  
  \vspace{-0.2cm} 
  \caption{(Left) \textbf{Visualizing Kinetic Regimes.} Comparison of different types of trajectories. 
  (Right) Distribution of Hit Counts in $R\to P\to W$: Hit multiple times but drifted away.}
  \label{fig:comparison}
  \vspace{-0.3cm}
\end{figure}

A distinctive advantage of \textbf{LatentRxnFlow} is that it enables \emph{trajectory based failure diagnosis}. Instead of observing only the final product prediction, we analyze the entire continuous latent trajectory $\{z_t\}_{t\in[0,1]}$ and define failure modes based on how decoded structures evolve over time. 
Concretely, we decode the latent state at 11 uniform time points $t \in \{0.0, 0.1, \dots, 1.0\}$. Let $\mathcal{G}_t=\mathcal{D}(z_t)$ be the decoded graph at time $t$, $\mathcal{G}_{gt}$ the ground-truth product, and $\mathcal{G}_r$ the reactant. We categorize each trajectory by the sequence $\{\mathcal{G}_t\}$:


\begin{enumerate}[leftmargin=*]
    \item \textbf{Optimal Convergence ($R \to P$).} The trajectory successfully evolves into the product basin. The final state at $t=1.0$ matches the ground truth ($\mathcal{G}_{1.0} = \mathcal{G}_{gt}$), shown as blue arrows in Fig. \ref{fig:comparison}.
    
    \item \textbf{Semantic Drift ($R \to W$).} The trajectory diverges to an incorrect basin. The final structure is chemically valid but wrong ($\mathcal{G}_{1.0} \neq \mathcal{G}_{gt}$), and the correct product \textit{never} appears in any intermediate step ($\forall t, \mathcal{G}_t \neq \mathcal{G}_{gt}$), shown as red arrows in Fig. \ref{fig:comparison}.
    
    \item \textbf{Kinetic Overshooting / Oscillation ($R \to P \to [W/R]$).} A "Hit@Path" event occurs. The trajectory correctly identifies the product at an intermediate step (e.g., $\mathcal{G}_{0.8} = \mathcal{G}_{gt}$), but fails to stabilize, subsequently drifting to an invalid state ($W$) or recoiling to the reactant ($R$) at $t=1.0$.
    
    \item \textbf{Reaction Failure ($R \to [R/W] \to R$).} The trajectory fails to escape the reactant's basin effectively. The final decoded structure reverts to the reactant ($\mathcal{G}_{1.0} = \mathcal{G}_{r}$), and the product is never visited. This includes both static stagnation and transient exploration of noise ($R \to W \to R$).
\end{enumerate}

\begin{table}[t]

    \centering
    \caption{\textbf{Inference Efficiency on NVIDIA RTX3090.} } \vspace{-0.1in}
    \label{tab:efficiency_comparison}
    \resizebox{0.5\textwidth}{!}{%
    \begin{tabular}{llccrcc}
    \toprule
    \textbf{Model}  & \textbf{Solver Scheme} & \textbf{ Steps} & \textbf{Latency} & \textbf{Speedup} & \textbf{Trajectory Nature} \\
    \midrule
    \textit{Baselines} \\
    FlowER (Large)  & Beam Search & $>$1000 & $\sim$64.0 s & 0.001$\times$ & Reaction Pathway   \\
    MEGAN  & Beam Search & $\sim$10 & 72.2 ms & 3.9$\times$ & Graph Edits \\
    Chemformer  & Beam Search & $\sim$50 & 280.8 ms & 1.0$\times$ & Not Available\\
    \midrule
    \textbf{Ours}  & \textbf{  ODE: RK4} & \textbf{10} & \textbf{8.5 ms} & \textbf{33.0$\times$} &  \textbf{Latent  Pathway}\\
    \textbf{Ours}  & \textbf{  ODE: RK4} & \textbf{100} & \textbf{8.6 ms} & \textbf{32.7$\times$}  & \textbf{Latent  Pathway} \\
    \bottomrule
    \end{tabular}%
    }
\end{table}

We analyzed the trajectory obtained on the USPTO-MIT test set. While our model achieves a high success rate (\textbf{$R \to P$: 90.2\%}), a breakdown of the 
failure cases reveals critical insights:
\begin{itemize}[leftmargin=*]
\item \textbf{Dominance of Semantic Drift ($R \to W$), 76.2\%} of  failure cases: the flow fails to reach the correct product basin .
\item \textbf{The Recoverable Margin ($R \to P \to W/R$), 17.5\%} of  failure cases: exhibiting Kinetic Overshooting / Oscillation, i.e., the vector field \textit{did} find the correct reaction path but "overshot" due to the lack of an explicit stopping mechanism.  These findings imply that nearly \textbf{1/6 of the model's errors are recoverable}.

\item \textbf{Low Reactivity ($R \to R/W \to R$), 6.2\%} of failure cases.  A small fraction of trajectories  fail to initiate a successful reaction, ending back at the reactant state. This low percentage suggests that the Flow Matching objective is highly effective at overcoming the activation barrier and encouraging kinetic transport compared to translation-based baselines.
\end{itemize}
 



On closer inspection of the $R \to P \to [W/R]$ cases, we find a surprising pattern: many trajectories enter the correct semantic region for $\ge 5$ integration steps (i.e., $>50\%$ of sampled trajectories) but drift away at the final step, as shown in Fig. \ref{fig:comparison}. This \emph{terminal instability} suggests the vector field can reach the target manifold but does not create a stable attractor ( a "sink" where $v(z) \to 0$) so the solver may oscillate or overshoot at the end. As a result, greedy ``take-the-last-step'' decoding is suboptimal; trajectory-aware decoding (e.g., stability monitoring or early stopping) could recover these near-miss correct predictions without retraining.


\vspace{-0.1in}
\subsection{Gated Inference: A Proof-of-Concept} \label{sec:gate_inference}
Motivated by the above failure diagnosis, we propose a simple \emph{trajectory intervention} at inference time as a proof-of-concept for error correction. The key idea is to detect \emph{terminal instability} and, when triggered, replace the final prediction with a more stable intermediate state in the trajectory.  Given a latent trajectory $\{z_t\}_{t=0}^1$ and its decoded structures $\{\mathcal{G}_t\}$, we  define
\textbf{two trajectory features}:
\begin{itemize}[leftmargin=*]
    \item \textbf{Terminal speed} $v_{\text{end}}$: the average velocity magnitude near the endpoint ($t\!\to\!1$). A large $v_{\text{end}}$ indicates non-convergence to a stationary basin (e.g., overshooting cases).
    \item \textbf{Structural dwell time} $\tau_{\text{stable}}$: the number of consecutive steps for which the decoded structure remains invariant. A small $\tau_{\text{stable}}$ suggests the prediction is transient and chemically unstable.
\end{itemize}


We reject the final prediction $\mathcal{G}_{1}$ if the trajectory exhibits
(i) RDKit parsing invalidity or an output identical to the reactant, and (ii) high terminal speed ($v_{\text{end}}>\delta_v$) together with low dwell time ($\tau_{\text{stable}}<\delta_\tau$). In these cases, we \emph{rewind} to the most stable intermediate state $\mathcal{G}_{\text{best}}$.
Empirically, this simple geometric gating recovers \textbf{18.93\%} of the $R \to P \to W$ failures and \textbf{84.38\%} of the $R\to P\to R$ failures. Overall, as described in Table ~\ref{tab:intervention}, 151 reactions are recovered from wrong to right, which is $\sim21.8\%$ of the recoverable failures, and $\sim3.84\%$ of all the failures in USPTO-MIT test set. (See Appendix algorithm ~\ref{alg:tagi_combined} for  details of the gated inference and Table ~\ref{tab:gated_performance} for detailed recovery rate grouped by fail modes).


The success of such a simple solution confirms that  \textbf{latent trajectories contain rich, high-fidelity signals} about prediction correctness. This ``free lunch'' from the ODE trajectory offers a promising direction for future work, where parameterized ``trajectory critics'' could be trained to further boost reliability.

\vspace{-0.1in}
\subsection{Geometric Analysis of Uncertainty}
\label{sec:geometric}
Another distinctive feature of our method is that the \textbf{latent geometry of the inferred trajectory} provides a lightweight uncertainty signal. To study this effect, we define a set of geometric descriptors from the discrete latent trajectory $\mathcal{T}=\{\mathbf{z}_t\}_{t=0}^{T}$, where $\mathbf{z}_t\in\mathbb{R}^d$ is the graph-pooled molecular representation at integration step $t$. We denote the instantaneous displacement (velocity) by $\mathbf{v}_t=\mathbf{z}_{t+1}-\mathbf{z}_t$.


\begin{enumerate}[leftmargin=*]
    \item \textbf{Path Inefficiency ($\eta$):} 
    Derived from the ratio of the cumulative path length to the Euclidean displacement, this metric quantifies the deviation from the optimal transport geodesic:
    \begin{equation}
        \eta = \frac{\sum_{t=0}^{T-1} \|\mathbf{v}_t\|_2}{\|\mathbf{z}_T - \mathbf{z}_0\|_2 + \epsilon},
    \end{equation}
    where $\epsilon$>0 for numerical stability.  An $\eta \approx 1$ indicates an ideal linear transport, while $\eta \gg 1$ implies a tortuous path indicative of high transport cost and manifold irregularity.

    \item \textbf{Mean Curvature ($\kappa$):} 
    To measure local flow fluctuations, we approximate the trajectory curvature using the mean norm of the discrete second-order differences (local acceleration):
    \begin{equation}
        \kappa = \frac{1}{T-2} \sum_{t=1}^{T-1} \|\mathbf{z}_{t+1} - 2\mathbf{z}_t + \mathbf{z}_{t-1}\|_2
    \end{equation}
    High curvature ($\kappa$) suggests that the vector field undergoes rapid directional changes, reflecting a rugged energy landscape or unstable solver dynamics.

    \item \textbf{Minimum Alignment ($\alpha_{min}$):} 
    We assess the monotonicity of the trajectory by calculating the cosine similarity between the local velocity $\mathbf{v}_t$ and the global displacement $\mathbf{v}_{global} = \mathbf{z}_T - \mathbf{z}_0$. We report the \emph{minimum alignment}:
    \begin{equation}
        \alpha_{min} = \min_{t \in [0, T-1]} \left( \frac{\mathbf{v}_t \cdot \mathbf{v}_{global}}{\|\mathbf{v}_t\|_2 \|\mathbf{v}_{global}\|_2} \right)
    \end{equation}
    A low $\alpha_{min}$ reveals "bottleneck" regions where the solver traverses orthogonally or backward relative to the target to circumnavigate latent barriers.

    \item \textbf{Latent Kinetic Energy ($\mathcal{K}$):} 
    Reflecting the magnitude of latent evolution, we define the mean kinetic energy as the average step size:
    \begin{equation}
        \mathcal{K} = \frac{1}{T} \sum_{t=0}^{T-1} \|\mathbf{v}_t\|_2
    \end{equation}
    Sudden spikes or vanishing values in $\mathcal{K}$ serve as indicators for kinetic instability or mode collapse, respectively.
\end{enumerate}

\begin{figure}[t]
    \centering
    \includegraphics[width=0.9\linewidth]{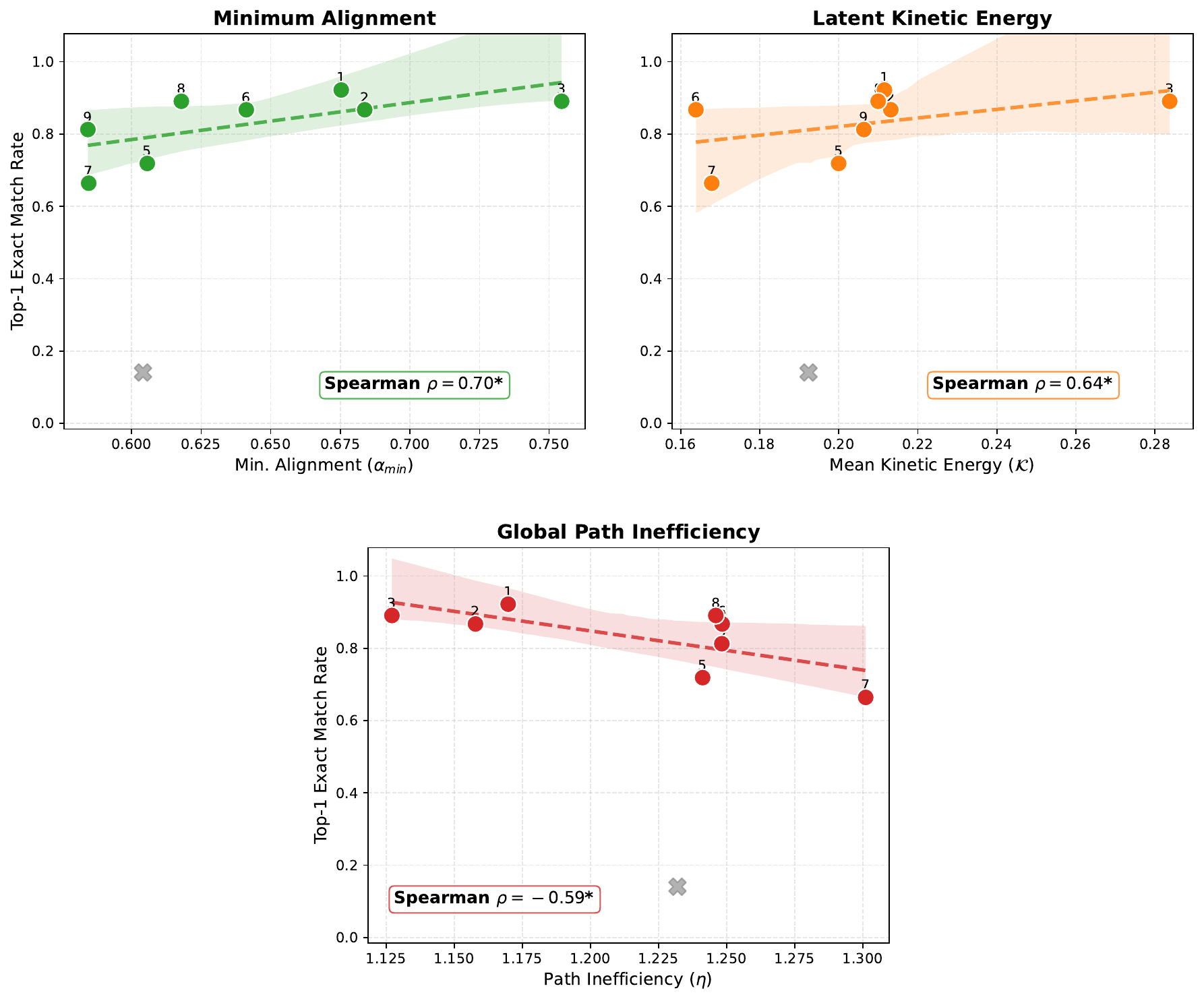}
    \caption{ Prediction accuracy vs. trajectory geometric descriptors, measured by mean value of $\alpha_{\min}$,  $\mathcal{K}$, and   $\eta$, across   reaction classes 1--9.  Straight lines show linear fits; shaded bands indicate 95\% confidence intervals.
    \textit{Note:} Reaction Class 4 (marked as gray `$\times$') with 
    very low accuracy ($14.1\%$) is treated as an outlier and excluded from the regression and the calculation of Spearman's rank correlation coefficient.  } 
    \label{fig:uncertainty} \vspace{-0.1in}
\end{figure}

\begin{table*}[t]
    \centering
    \caption{\textbf{Comparisons on USPTO-MIT.} ``$\dagger$'' denotes results copied from original papers. 
    $^\ddagger$   denotes results trained with different data (USPTO-FULL--derived mechanism set) and is \textbf{not directly comparable}; reported for reference only.}
     \vspace{-0.1in}
    \label{tab:main_results}
    \resizebox{0.7\textwidth}{!}{
    \begin{tabular}{llccccccccc}
        \toprule
        & & & \multicolumn{5}{c}{\textbf{Prediction Accuracy (\%)}} & \\
        \cmidrule(lr){4-8}
        \textbf{Model} & \textbf{Generative Paradigm} & \textbf{Search Strategy} & \textbf{Top-1} & \textbf{Top-2} & \textbf{Top-3} & \textbf{Top-5} & \textbf{Top-10} \\
        \midrule
        \multicolumn{9}{l}{\textit{Sequence-based (SMILES)}} \\
        Chemformer Large$^\dagger$ & Sequence (Autoregressive) & Beam Search & 91.3 & - & - & 93.7 & 94.0 \\
        Chemformer$^\dagger$ & Sequence (Autoregressive) & Beam Search & 90.9 & - & - & 93.8 & 94.1 \\
        Molecular Transformer$^\dagger$ & Sequence (Autoregressive) & Beam Search & 88.7 & 92.1 & 93.1 & 94.2 & 94.9 \\
        Graph2SMILES (D-GCN)$^\dagger$ & Hybrid (Autoregressive) & Beam Search & 90.3 & - & 94.0 & 94.6 & 95.2 \\
        \midrule
        \multicolumn{9}{l}{\textit{Graph-based \& Hybrid (Explicit Discrete Trajectories)}} \\
    GTPN$^\dagger$ & Graph \textbf{(Step-wise Edit)} & Beam Search & 83.2 & - & 86.0 & 86.5 & - \\
    MEGAN$^\dagger$ & Graph \textbf{(Step-wise Edit)} & Beam Search & 86.3 & 90.3 & 92.4 & 94.0 & 95.4 \\
    \color{gray}FlowER  (on USPTO-Full)    $^\dagger$$^\ddagger$ & \color{gray}Graph (Mechanism Pathway) & \color{gray}Beam Search & \textit{\color{gray}92.5} & \textit{\color{gray}96.9} & \textit{\color{gray}98.2} & \textit{\color{gray}98.6} & \textit{\color{gray}98.7} \\
        
        \midrule
        \multicolumn{9}{l}{\textit{Non-Autoregressive Baselines}} \\
        WLDN$^\dagger$ & Combinatorial (Template) & Enum-Filter & 79.6 & - & 87.7 & 89.2 & - \\
        
        NERF $^\dagger$ & Latent \textbf{(One-shot VAE)} & Stochastic & 90.7 & 92.3 & 93.3 & 93.7 & - \\
        \midrule
        \multicolumn{9}{l}{\textit{\textbf{LatentRxnFlow (Ours) - Latent Flow Matching}}} \\
        \rowcolor{gray!10} Ours (w/o Flow) & Latent (One-shot GAE) & Deterministic & 89.6	&90.9	&91.8	&92.5	&92.8 \\
        \rowcolor{gray!10} Ours (w/o FiLM Conditional Injection) & Latent \textbf{(ODE Flow)} & Stochastic & 90.1	&91.0	&91.7	&92.1	&92.5 \\
        \rowcolor{gray!10} \textbf{Ours (full)} & Latent \textbf{(ODE Flow)} & Stochastic & 90.2 & 91.8 & 92.9 & 94.0 & 94.8 \\
        \bottomrule
        \bottomrule
    \end{tabular}
    }
\end{table*}

To investigate whether latent trajectory geometry can serve as an \emph{unsupervised proxy for prediction confidence}, we compute our geometric descriptors and correlate them with Top-1 accuracy across reaction classes 1--9 in USPTO-50k. For efficiency, we subsample the first 128 reactions from each class.  
Fig. \ref{fig:uncertainty}   reveals that successful reaction prediction requires both consistent directionality and sufficient kinetic magnitude.
Specifically, 
\begin{itemize}[leftmargin=*]
    \item \textbf{Bottleneck Alignment ($\alpha_{\min}$):} Strongest signal (Spearman $\rho=0.70$). High accuracy occurs when even the least-aligned step remains directed toward the target, suggesting bottlenecks dominate outcomes.
    \item \textbf{Decisiveness ($\mathcal{K}$):} Strong positive correlation ($\rho=0.64$). Successful reactions exhibit sustained latent motion, whereas low $\mathcal{K}$ often indicates stagnation near the reactant basin.
    \item \textbf{Transport Efficiency ($\eta$):} Moderate negative correlation ($\rho=-0.59$). More tortuous trajectories are riskier, but less predictive than alignment and decisiveness.
\end{itemize}
Intriguingly, local curvature ($\kappa$) shows negligible correlation ($\rho \approx -0.07$), indicating that high-frequency fluctuations are tolerable as long as the global direction ($\alpha$) and momentum ($\mathcal{K}$) are maintained.
We hypothesize that \textbf{high curvature is a functional necessity} for bypassing high-energy barriers in the latent landscape. 
When the model encounters an unstable  semantic region, it must execute a sharp maneuver to maintain the feasibility of the reaction path.

For each reaction class, we compute the median geometric descriptors over 128 sampled trajectories, presented in Table~\ref{tab:geo_stats} (in Appendix \ref{app:results}). For an intuitive comparison, we focus on two representative classes with contrasting confidence profiles: \textbf{1) High-confidence (Class 3, C--C bond formation)} with high accuracy, exhibits the most confident dynamics: highest kinetic energy ($\mathcal{K}$) and flow consistency ($\alpha_{\min}$), with minimal path inefficiency ($\eta$). This class serves as a baseline for "easy" geometric transport. \textbf{2) Low-confidence (Class 7, reduction)} with  lower accuracy,  characterized by  severe bottlenecks (lowest $\alpha_{\min}$) and high trajectory tortuosity (highest $\eta$),  represents the "hard" regime where the model faces significant semantic uncertainty.



\begin{figure}[t]
    \centering
    \includegraphics[width=\linewidth]{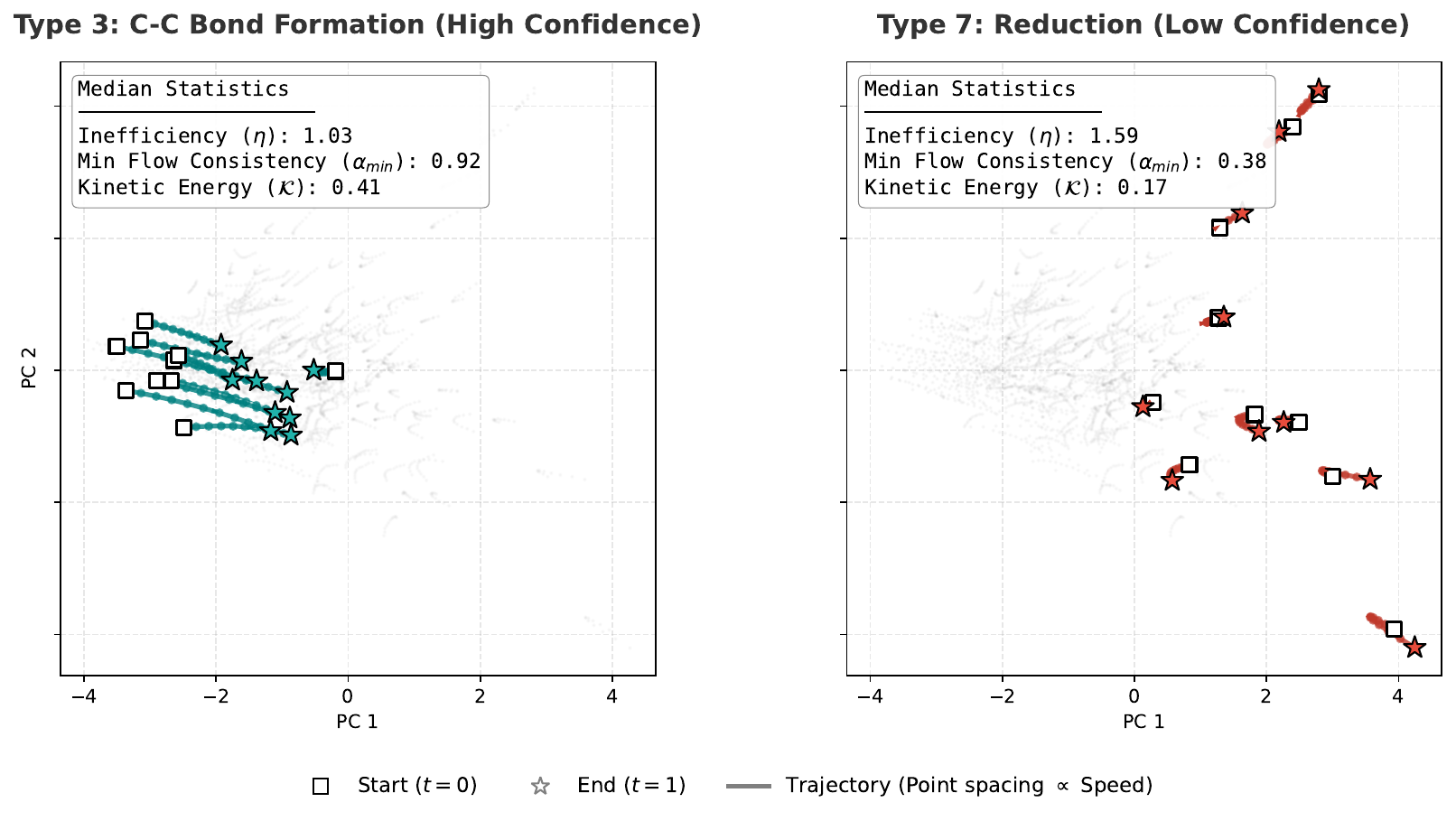}
    
    \caption{Visualizing latent dynamics under high vs. low confidence (Type 3 C-C Coupling vs. Type 7 Reduction).
    \underline{Left}: nearly linear, decisive paths with high kinetic energy ($\mathcal{K} \approx 0.41$) and flow alignment ($\alpha_{min} \approx 0.92$);
    \underline{Right}: significant tortuosity and "stagnation", characterized by high inefficiency ($\eta \approx 1.59$) and low min. alignment ($\alpha_{min} \approx 0.38$). }
    \vspace{-0.3in}
    \label{fig:traj_viz}
\end{figure}

Figure~\ref{fig:traj_viz} illustrates how latent-trajectory geometry tracks prediction confidence by projecting 10 samples via PCA. High-confidence trajectories (Type 3: C--C coupling) move in a coherent, decisive direction, with strong alignment and sustained latent motion. In contrast, low-confidence trajectories (Type 7: reduction) exhibit detours and dense clusters, indicating bottlenecks and stagnation. These trends match our quantitative results: higher $\alpha_{\min}$ and $\mathcal{K}$ correlate with success, while larger $\eta$ signals uncertainty, supporting trajectory geometry as a supervision-free confidence cue for trajectory-aware decoding.

\section{Conclusion}
\label{sec:conclusion} 
We presented a framework for reaction prediction using latent Conditional Flow Matching, establishing a taxonomy of  failure modes of reaction trajectories, demonstrating that latent dynamics are intrinsic indicators of success. To address the identified "kinetic overshooting," we introduced Trajectory-Aware Gated Inference, which recovers unstable predictions without retraining. Future work will extend this to \textbf{guided generation}, where a learnable critic actively steers the neural ODE solver toward chemically valid attractors, leading to continuous, proactive, and flexible  guidance.

\bibliographystyle{ACM-Reference-Format}

\appendix
\section{Implementation Details and More  Results} 

\subsection{Implementation Details on Main Results}
\label{app:impl_details}
\textbf{Implementation Details.} The GAE backbone employs a 6-layer Transformer encoder and 6-layer Transformer decoder with hidden dimension $d=256$ and 8 attention heads. The condition vector has dimension $d_c=640$ (128-bit frequent agent vector + 512-bit fingerprint embedding). We train the model for 500 epochs with a batch size of 256 and a learning rate of $3 \times 10^{-4}$.

\noindent
\textbf{Solver Settings} To balance training efficiency and inference precision, we employ a "Solver-Mismatch" strategy. During training, we use the Heun solver with $N=5$ steps (fast approximation); during inference, we switch to the Runge-Kutta 4 (RK4) solver with $N=20$ steps and $\sigma=0.05$ noise scaling. This demonstrates the robustness of the learned vector field.
\subsection{Endpoint Performance: Overall Analysis}
\label{app:overall-res}
\begin{itemize}

\item Ours vs. Discrete Trajectory Models: LatentRxnFlow significantly outperforms previous graph generative models like MEGAN (86.3\%) and GTPN (83.2\%). This confirms that modeling reactions as continuous flow in latent space is more effective than discrete autoregressive edits.

\item Ours vs. Sequence-based Translation Models: We match the performance of Graph 2SMILES (90.3\%), the current SOTA. Crucially, unlike translation models that act as "black boxes," LatentRxnFlow offers trajectory-based interpretability and uncertainty quantification (discussed below).

\item Ours vs. non-autoregressive Latent Baselines: Our model consistently outperforms deterministic latent baselines such as WLDN (79.6\%) and ours (w/o Flow). While the stochastic NERF achieves a marginally higher Top-1 accuracy (90.7\% vs 90.2\%), our FiLM variant surpasses it in Top-5 accuracy (94.0\% vs. 93.7\%), demonstrating that our flow matching objective provides a more robust coverage of the reaction distribution without relying on the specific priors of a VAE.

\item Condition Injection via Input vs via CFM Module: We compare injecting conditions via simple input concatenation (Base/RRCC) versus injecting them into the flow dynamics (via FiLM). While Input Injection achieves a respectable Top-1 of 90.1\%, the Module-based injection significantly boosts Top-k performance (e.g., FiLM achieves 94.8\% Top-10 vs. Base's 92.5\%). This suggests that structurally modulating the vector field (Condition Injection) is superior to static input concatenation for guiding the generation of diverse valid products.

\item FiLM vs. ControlNet: Comparing our injection mechanisms, the Residual FiLM variant outperforms the ControlNet-style gating by a margin of $\sim$0.5\%, suggesting that modulating the global energy landscape (FiLM) is more suitable for reaction conditions than spatial feature injection.

\item Fingerprint Set vs. Bit And: Comparing the two conditioning formats, the "Fingerprint Set" representation yields significantly better long-tail performance than the "Bit And" operation. While "Bit And" has a slight edge in Top-1 (90.3\% vs 90.2\%), "Fingerprint Set" dominates in Top-10 accuracy (94.8\% vs 93.7\%), indicating that preserving the discrete set structure of conditions provides richer information for the model than a compressed bitwise representation.

\item \textit{Ours vs. FlowER: We include FlowER (grayed out in Table~\ref{tab:main_results}) solely as a reference for performance scale at the high-data regime. Note that FlowER was trained and evaluated on the massive USPTO-Full dataset ($\approx$1.1M reactions with 1.4M elementary steps), whereas our model and other baselines (MEGAN, GTPN, Graph2SMILES) are evaluated on the standard USPTO-MIT benchmark. The performance gap is primarily attributed to this vast difference in data availability. Due to computational resource constraints (see Table~\ref{tab:efficiency_comparison} for efficiency analysis, we need 60s/sample*40000samples$\approx$ \textbf{667 GPU hrs (nearly one month)} on NVIDIA RTX3090  to run full test on USPTO-MIT ), retraining or strictly evaluating FlowER on the USPTO-MIT test set was not feasible in this study. In contrast, our model completes the same task in roughly 5 mins, demonstrating a $\sim8000 \times$ efficiency advantage in practical throughput}

\end{itemize}
\begin{table}[h]
\centering
\caption{\textbf{Ablation Study on Conditioning Mechanisms.} We analyze the impact of injection strategy, module architecture, and data representation format on prediction accuracy.}
\label{tab:ablation_all}
\resizebox{\linewidth}{!}{%
\begin{tabular}{lccccc}
\toprule
 & \multicolumn{5}{c}{\textbf{Prediction Accuracy (\%)}} \\
\cmidrule(lr){2-6}
\textbf{Model Variant} & \textbf{Top-1} & \textbf{Top-2} & \textbf{Top-3} & \textbf{Top-5} & \textbf{Top-10} \\
\midrule
\multicolumn{6}{l}{\textit{1. Condition Injection Strategy}} \\
Input Concatenation (Base/RRCC) & 90.1 & 91.0 & 91.7 & 92.1 & 92.5 \\
Module Injection (FiLM + Set) & \textbf{90.2} & \textbf{91.8} & \textbf{92.9} & \textbf{94.0} & \textbf{94.8} \\
\midrule
\multicolumn{6}{l}{\textit{2. Module Architecture (Flow Modulation)}} \\
ControlNet Gating & \textbf{90.3} & 91.7 & 92.5 & 93.1 & 93.5 \\
Residual FiLM  & 90.2 & \textbf{91.8} & \textbf{92.9} & \textbf{94.0} & \textbf{94.8} \\
\midrule
\multicolumn{6}{l}{\textit{3. Condition Data Representation}} \\
Bitwise "AND" Operation & \textbf{90.3} & 91.8 & 92.6 & 93.2 & 93.7 \\
Fingerprint "Set" Structure & 90.2 & \textbf{91.8} & \textbf{92.9} & \textbf{94.0} & \textbf{94.8} \\
\bottomrule
\end{tabular}
}
\end{table}

\section{Efficiency Analysis} \label{app:Efficiency}
\subsubsection{Theoretical Complexity: The Quadratic-to-Linear Shift}A critical architectural decision is \textit{where} to inject the reaction conditions $C$ (e.g., reagents, catalysts). We compare two standard paradigms:\begin{itemize}\item \textbf{Input-Level Fusion (RRCC):} Concatenating condition tokens with reactant tokens $R$ at the encoder input.\item \textbf{Head-Level Modulation (Ours):} Encoding only reactants $R$ in the backbone and injecting $C$ into the flow head $v_\theta$ via FiLM.\end{itemize}Let $L_R$ be the sequence length of reactants and $L_C$ be the sequence length of conditions. Let $d$ be the hidden dimension of the Transformer backbone, and $H$ be the width of the Flow Head MLP.

\paragraph{Backbone Complexity Analysis.}The computational bottleneck of a Transformer encoder lies in the Self-Attention mechanism, which scales quadratically with sequence length.\begin{itemize}\item \textbf{Input-Level Fusion:} The effective sequence length is $L_{total} = L_R + L_C$. The attention complexity is $\mathcal{O}((L_R + L_C)^2 \cdot d)$.\item \textbf{Head-Level Modulation (Ours):} The backbone only processes reactants, so $L_{total} = L_R$. The complexity is $\mathcal{O}(L_R^2 \cdot d)$.\end{itemize}By decoupling conditions from the backbone, we eliminate the cross-term $2L_R L_C$ and the self-term $L_C^2$ from the heavy attention computation.

\paragraph{Flow Head Complexity with FiLM } 
In our proposed architecture, the condition vector $c$ modulates the flow field via Feature-wise Linear Modulation (FiLM). For a hidden layer $h$ in the MLP head, the modulation is defined as:\begin{equation}\text{FiLM}(h, c) = \gamma(c) \odot h + \beta(c)\end{equation}where $\gamma(\cdot)$ and $\beta(\cdot)$ are lightweight linear projections mapping $c$ to scale and shift parameters, and $\odot$ denotes element-wise multiplication.\begin{itemize}\item \textbf{Computational Cost:} The modulation introduces only two linear projections $\mathcal{O}(d_c \cdot H)$ and element-wise operations $\mathcal{O}(H)$.\item \textbf{Comparison:} Unlike Input-Level Fusion which forces the backbone to model explicit interactions between every condition token and reactant token (attention map of size $(L_R+L_C)^2$), FiLM enforces a global, efficient conditioning. The complexity is strictly linear with respect to condition dimension $d_c$, i.e., $\mathcal{O}(d_c)$.\end{itemize}\paragraph{The Efficiency Trade-off: Quadratic vs. Linear.}Our design fundamentally shifts the burden of condition processing:\begin{itemize}\item \textbf{Backbone (Removal of Quadratic Term):} By removing $L_C$ tokens from the Transformer input, we save $\mathcal{O}(2 L_R L_C + L_C^2)$ attention operations per layer. Given that reaction conditions (reagents, catalysts, solvents) often require long textual descriptions (large $L_C$), this reduction is significant.\item \textbf{Head (Addition of Linear Term):} We introduce a negligible overhead of $\mathcal{O}(d_c)$ in the shallow MLP head via FiLM.\end{itemize}This architectural decoupling ensures that the heavy Transformer backbone focuses solely on the complex molecular graph reasoning ($\mathcal{O}(L_R^2)$), while the static condition information controls the vector field dynamics through efficient, low-rank modulation.

\paragraph{Training Throughput Analysis.}
We empirically validate the efficiency gain of our architectural decoupling by comparing the per-epoch training time with same hyperparameters, on USPTO-MIT 480k training split, on the same hardware (4 * NVIDIA A10 24G ):
\begin{itemize}
    \item \textbf{Input Fusion (RRCC):} Requires \textbf{$\approx$ 400s/epoch}. The inclusion of lengthy condition tokens (reagents, catalysts) significantly inflates the effective sequence length, triggering the quadratic complexity bottleneck of the Transformer's self-attention mechanism ($\mathcal{O}((L_R+L_C)^2)$).
    \item \textbf{Head Modulation (Ours):} Requires only \textbf{$\approx $180s/epoch}, achieving a \textbf{2.2$\times$ speedup}. By offloading condition processing to the linear FiLM head ($\mathcal{O}(d_c)$), we bypass the redundant attention computations between static condition tokens and dynamic reactant tokens.
\end{itemize}
This acceleration not only reduces the carbon footprint of training but also enables faster experimental iteration cycles, proving that modeling the \textit{Reaction-Condition} interaction via vector field modulation is both semantically superior and computationally efficient.


\subsubsection{Inference Efficiency}
\paragraph{Experimental Setup}
To rigorously evaluate inference speed, we benchmarked all models on a single NVIDIA RTX3090 GPU. For MEGAN, Chemformer, and Ours, we measured the average latency over the first 5,000 samples of the USPTO-MIT test set to ensure statistical stability. For FlowER, due to its extremely high computational cost (requiring $>1000$ sampling steps per reaction), we restricted the evaluation to the first 64 samples to obtain a feasible estimate of its runtime.

\paragraph{Results and Analysis}
Table \ref{tab:efficiency_comparison} summarizes the inference efficiency across different solver schemes. Our method demonstrates a significant performance advantage:
\begin{itemize}\item \textbf{Extreme Speedup:} Our model achieves an average latency of $\sim$8.5 ms per sample, representing a 33$\times$ speedup over the autoregressive Transformer baseline (Chemformer) and an 8.5$\times$ speedup over the graph-edit model (MEGAN). Compared to FlowER, which operates on the explicit reaction pathway ($\sim$64.0 s), our method is orders of magnitude faster.
\item \textbf{Decoupling Dynamics from Decoding (Key Benefit):} A critical observation is that increasing the ODE integration steps from 10 to 100 results in negligible latency overhead (8.5 ms vs. 8.6 ms). This phenomenon strongly validates the efficiency of our **semi-black box** design. Unlike autoregressive models (Chemformer) that run the heavy network for every token ($\sim$50 times), or pathway models (FlowER) that manipulate heavy graph structures at every step, our trajectory integration occurs in a \textit{low-dimensional, computationally lightweight latent space}. The primary computational cost is incurred only once during the final decoding from latent vector to molecular graph. Consequently, we can increase the solver precision (steps) without compromising inference speed.
\end{itemize}
\subsubsection{Lateral Comparison: The Cost of Precision is Negligible}We benchmark the inference latency against state-of-the-art baselines. For LatentRxnFlow, we employ the \textbf{Runge-Kutta 4 (RK4)} solver to ensure high-precision integration of the reaction trajectory.\paragraph{The "Fixed-Cost" Phenomenon.}As shown in Table~\ref{tab:efficiency_comparison}, LatentRxnFlow operates in a unique \textbf{fixed-cost dominated regime}.Despite RK4 requiring 4 network evaluations per step, increasing the integration steps from 10 to 100 incurs almost \textbf{no latency penalty} (8.5 ms vs. 8.6 ms).This indicates that the computational bottleneck lies entirely in the one-time graph encoding/decoding overhead ($\mathcal{O}(1)$ w.r.t steps), while the continuous flow integration in the compact latent space is computationally "free."

\paragraph{Paradigm Shift in Efficiency.}Even with the robust RK4 solver, LatentRxnFlow achieves a throughput of $\sim$117 molecules/sec, delivering a 33$\times$ speedup over Chemformer and an 8.5$\times$ speedup over MEGAN. This result challenges the common belief that ODE-based sampling is slow. By decoupling the generative resolution (Steps) from the structural complexity (Graph/Sequence length), LatentRxnFlow achieves real-time inference without sacrificing physical consistency.

\section{Inference-Time Traject-Aware Gated Inference  Algorithm}

Our analysis reveals that kinetic instability (e.g., overshooting) leaves distinct signatures: high residual velocity ($\bar{v}_{end}$) and transient semantic dwelling (short $L_{plateau}$). Based on this, we design a gating mechanism (Algorithm \ref{alg:tagi_combined}) to dynamically select between the final prediction $\mathcal{M}_T$ and an intermediate candidate $\mathcal{M}_{t^*}$. The gate triggers a "rescue" operation under two conditions: 
\begin{itemize}
    \item Trivial Failure: If the final output is chemically invalid or identical to the reactant (Identity Map), we fallback to the best intermediate valid structure. 
    \item Kinetic Instability: If the trajectory exhibits high end-speed ($\bar{v}_{end} > 0.15$) and fails to form a stable semantic plateau ($L_{plateau} < 3$), we infer that the solver has overshot the target. In this case, we retrieve the prediction from the most stable intermediate step $t^*$.
\end{itemize}
 For stable trajectories (low velocity, long plateau), the gate remains closed, preserving the final output to minimize false corrections (R$\to$W).


\begin{algorithm*}[t]
\caption{Trajectory-Aware Gated Inference (TAGI)}
\label{alg:tagi_combined}
\DontPrintSemicolon

\SetKwInOut{Input}{Input}
\SetKwInOut{Output}{Output}

\Input{Latent trajectory $\mathbf{z}_{0:T}$, Decoded molecules $\mathcal{M}_{0:T}$, Reactant $\mathcal{R}$, Thresholds $\tau_{speed}=0.15, \tau_{plateau}=3, \lambda$}
\Output{Final Prediction $\hat{y}$}

\BlankLine
\tcp{\textbf{Phase I: Trajectory Stability Analysis}}
\tcp{1. Compute instantaneous velocity profile}
Initialize velocities $\mathbf{v} \leftarrow []$\;
\For{$t \leftarrow 1$ \KwTo $T$}{
    $v_t \leftarrow \|\mathbf{z}_t - \mathbf{z}_{t-1}\|_2$\;
    Append $v_t$ to $\mathbf{v}$\;
}
\tcp{Compute mean velocity over the last 30\% of trajectory}
$T_{end} \leftarrow \lfloor 0.7 \times T \rfloor$\;
$\bar{v}_{end} \leftarrow \text{Mean}(\mathbf{v}[T_{end}:T])$\;

\BlankLine
\tcp{2. Identify semantic stability (Plateau backwards from T)}
$L_{plateau} \leftarrow 0$\;
\For{$t \leftarrow T$ \KwTo $1$}{
    \If{$\text{Sim}(\mathcal{M}_t, \mathcal{M}_{t-1}) \ge 0.995$}{
        $L_{plateau} \leftarrow L_{plateau} + 1$\;
    }
    \Else{
        \textbf{break}\;
    }
}

\BlankLine
\tcp{3. Select best intermediate candidate $t^*$}
Initialize candidates $\mathcal{C} \leftarrow []$\;
\For{$t \leftarrow 1$ \KwTo $T$}{
    \If{$\mathcal{M}_t$ is valid \textbf{and} $\mathcal{M}_t \neq \mathcal{R}$}{
        Add $t$ to $\mathcal{C}$\;
    }
}

\eIf{$\mathcal{C}$ is empty}{
    $t^* \leftarrow T$ \tcp*{Fallback to final step}
}{
    \tcp{Maximize plateau, minimize velocity}
    $t^* \leftarrow \operatorname*{arg\,max}_{t \in \mathcal{C}} (\text{PlateauLength}(t) - \lambda v_t)$\;
}

\BlankLine
\BlankLine
\tcp{\textbf{Phase II: Gated Inference Decision}}
$\hat{y}_{final} \leftarrow \mathcal{M}_T$\;
$\hat{y}_{best} \leftarrow \mathcal{M}_{t^*}$\;

\BlankLine
\tcp{Condition 1: Trivial Failure (Invalid or Identity)}
\If{$\hat{y}_{final}$ is Invalid \textbf{or} $\hat{y}_{final} == \mathcal{R}$}{
    \Return $\hat{y}_{best}$ \tcp*{Immediate Rescue}
}

\BlankLine
\tcp{Condition 2: Kinetic Instability (Overshooting)}
\If{$\bar{v}_{end} > \tau_{speed}$ \textbf{and} $L_{plateau} < \tau_{plateau}$}{
    \tcp{High residual velocity + Short plateau $\to$ Overshot}
    \Return $\hat{y}_{best}$ \;
}

\BlankLine
\tcp{Condition 3: Stable Trajectory (Default)}
\Return $\hat{y}_{final}$ \tcp*{Trust the solver}

\end{algorithm*}

\section{Additional Experimental Results}
\label{app:results}

\subsection{Detailed Geometric Statistics by Reaction Class}
In this section, we provide a granular analysis of the geometric properties associated with  reaction types 1--9 in the USPTO-50k dataset. To understand the topological complexity of modeling different transformations, we computed the median values of four geometric descriptors: Kinetic Energy ($\mathcal{K}$), Curvature ($\kappa$), Minimum Alignment ($\alpha_{\min}$), and Path Inefficiency ($\eta$). The results are summarized in Table \ref{tab:geo_stats}.

We observe significant heterogeneity in the geometric behavior of different reaction classes, which correlates with the underlying chemical complexity:
\begin{itemize}\item \textbf{High-Confidence Dynamics (Type 3):} The \textit{C-C Bond Formation} class exhibits the most robust geometric signatures. It achieves the highest Kinetic Energy ($\mathcal{K}=0.265$) and Flow Consistency ($\alpha_{\min}=0.770$), alongside the lowest Inefficiency ($\eta=1.102$). This suggests that the model learns to represent C-C bond formation as a direct, high-momentum trajectory through the latent manifold, characterized by minimal bottlenecks.
\item \textbf{Complex Regimes (Type 7):} In contrast, \textit{Reductions} appear to be the most geometrically challenging category. This class is characterized by the lowest flow consistency ($\alpha_{\min}=0.607$) and the highest path inefficiency ($\eta=1.293$). The low kinetic momentum ($\mathcal{K}=0.165$) and high tortuosity indicate that the model struggles to find a linear interpolation for reduction reactions, likely due to the diverse and drastic structural changes often involved in these transformations.

\item \textbf{General Trends:} Other classes, such as \textit{Protections} (Type 5) and \textit{Deprotections} (Type 6), also show elevated inefficiency levels compared to simple alkylations. This geometric analysis provides quantitative evidence that certain reaction types impose greater topological constraints on the generative process than others.
\end{itemize}
\begin{table}[t]
    \centering
    \caption{\textbf{Geometric Statistics by Reaction Class (Median Values).} 
    We report the median geometric descriptors for the 9 reaction classes in USPTO-50k.
    \textbf{Type 3 (C-C Bond Formation)} exhibits the most confident dynamics: highest kinetic energy ($\mathcal{K}$) and flow consistency ($\alpha_{\min}$), with minimal path inefficiency ($\eta$).
    In contrast, \textbf{Type 7 (Reductions)} represents the most difficult regime, characterized by low momentum, severe bottlenecks (lowest $\alpha_{\min}$), and high trajectory tortuosity (highest $\eta$).}
    \label{tab:geo_stats}
    \setlength{\tabcolsep}{5pt} 
    \resizebox{\linewidth}{!}{
    \begin{tabular}{lcccc}
        \toprule
        \textbf{Reaction Type} & \textbf{Kinetic Energy} & \textbf{Curvature} & \textbf{Min. Flow Consistency} & \textbf{Inefficiency} \\
        (ID) & ($\mathcal{K}$) $\uparrow$ & ($\kappa$) $\downarrow$ & ($\alpha_{\min}$) $\uparrow$ & ($\eta$) $\downarrow$ \\
        \midrule
        1. Heteroatom Alkylation & 0.206 & 0.037 & 0.700 & 1.154 \\
        2. Acylation & 0.211 & 0.038 & 0.689 & 1.148 \\
        \textbf{3. C-C Bond Formation} & \textbf{0.265} & 0.041 & \textbf{0.770} & \textbf{1.102} \\
        4. Heterocycle Formation & 0.191 & 0.040 & 0.613 & 1.217 \\
        5. Protections & 0.199 & 0.043 & 0.622 & 1.238 \\
        6. Deprotections & 0.156 & 0.035 & 0.652 & 1.234 \\
        \textbf{7. Reductions} & 0.165 & 0.039 & \textbf{0.607} & \textbf{1.293} \\
        8. Oxidations & 0.197 & 0.041 & 0.667 & 1.179 \\
        9. FGIs & 0.199 & 0.042 & 0.614 & 1.225 \\
        \bottomrule
    \end{tabular}
    }
\end{table}

\subsection{Detailed Analysis of Gated Inference Dynamics}
In this section, we provide a granular analysis of the inference trajectories and the impact of our Gated Inference mechanism (Table~\ref{tab:gated_performance} and Table~\ref{tab:intervention}). We categorize the generation process based on the temporal evolution of the predicted product state and evaluate how the gating module interacts with different error modes.

As shown in Table~\ref{tab:gated_performance}, our Gated mechanism is particularly effective at addressing temporal instability: High Rescue Rate for Oscillation (84.4\%): The critic module successfully identifies the high-confidence plateau associated with the state $P$ before the model reverts to $R$. This confirms that $R \to P \to R$ errors are primarily due to rigid integration steps rather than model incapacity. Moderate Rescue for Overshooting (18.9\%): For $R \to P \to W$ trajectories, the critic can occasionally intervene when the transition from $P$ to $W$ involves a perceptible drop in likelihood or energy stability.Minimal Corruption on Stable Samples (0.23\%): Crucially, the false positive rate (triggering early stops on already correct trajectories, turning $P \to W$) is negligible. This indicates that the gating threshold is robust and does not disrupt well-formed reaction pathways.

Table~\ref{tab:intervention} quantifies the aggregate impact of the Gated Critic compared to the standard fixed-step inference and the theoretical upper bound (Oracle).\begin{itemize}\item \textbf{Oracle Potential (Best-of-Path):} We define the "Best-of-Path" accuracy (91.93\%) as the performance achieved if an Oracle perfectly selects the correct state $P$ whenever it appears in the trajectory. The gap between Standard Inference (90.17\%) and the Oracle highlights the potential gain from dynamic stopping.\item \textbf{Cost-Benefit Analysis:} Our Gated mechanism achieves a net accuracy gain (+0.16\%) by successfully rescuing 151 failure cases (flipping $W \to R$) while only corrupting 82 correct cases.\item \textbf{Efficiency:} The mechanism is lightweight, triggering an intervention in only 9.3\% of the test cases. This selective activation ensures that the computational overhead is minimized, as the majority of trajectories are allowed to proceed to full integration without interference.\end{itemize}

The Gated Inference mechanism effectively acts as a conservative "safety brake," recovering correct solutions from unstable trajectories that would otherwise be lost due to over-generation or reversion, bridging the gap between standard inference and the theoretical Oracle performance.
\begin{table}[t]
    \centering
    \caption{\textbf{Gated Inference Performance by Trajectory Type.} 
    The proposed gating mechanism effectively "rescues" unstable trajectories with minimal cost to established correct predictions.
    \textbf{$R \to P \to R$ (Oscillation)} sees a massive recovery rate ($84.4\%$), confirming that these errors are purely due to lack of stopping.
    \textbf{$R \to P \to W$ (Overshooting)} sees a moderate recovery ($18.9\%$).
    Crucially, for already correct samples (\textbf{Type E}), the corruption rate (R$\to$W) is negligible ($0.23\%$).}
    \label{tab:gated_performance}
    \setlength{\tabcolsep}{5pt}
    \resizebox{\linewidth}{!}{
    \begin{tabular}{lccccc}
        \toprule
        \textbf{Trajectory Type} & \textbf{Count} & \textbf{Trigger Rate} & \textbf{Rescue Rate} & \textbf{Corruption Rate} & \textbf{Net Gain} \\
        (Description) & ($N$) & ($\tau$) & (W $\to$ R) $\uparrow$ & (R $\to$ W) $\downarrow$ & ($\Delta$) \\
        \midrule
        \multicolumn{6}{l}{\textit{Failure Modes (Original Prediction was Wrong)}} \\
        $R\to R\to R$ (Stagnation) & 114 & 100.0\% & 0.0\% & - & 0 \\
        \textbf{$R \to P \to R$ (Oscillation)} & 32 & 100.0\% & \textbf{84.4\%} & - & +27 \\
        \textbf{$R \to P \to W$ (Overshooting)} & 655 & 26.0\% & \textbf{18.9\%} & - & +124 \\
         $R \to W \to R$  (Divergence) & 131 & 100.0\% & 0.0\% & - & 0 \\
        $R\to W$(Drift) & 2,995 & 13.1\% & 0.0\% & - & 0 \\
        \midrule
        \multicolumn{6}{l}{\textit{Success Modes (Original Prediction was Correct)}} \\
        \textbf{$R\to P$ (Stable)} & 36,057 & 8.2\% & - & \textbf{0.23\%} & -82 \\
        \bottomrule
    \end{tabular}
    }
\end{table}

\begin{table}[t]
\centering
\caption{\textbf{Rescue Performance via Trajectory Intervention.} By applying a lightweight critic (Gated mechanism) on overshooting / oscillation errors, we can stop the trajectory early when it hits a high-confidence plateau, converting failures into successes without retraining.}
\label{tab:intervention}
\resizebox{\linewidth}{!}{
\begin{tabular}{lcccc}
\toprule
\textbf{Method} & \textbf{Accuracy} & \textbf{Flip (W$\to$R)} & \textbf{Flip (R$\to$W)} & \textbf{Trigger Rate} \\
\midrule
Standard Inference & 90.17\% & - & - & - \\
Best-of-Path (Oracle) & 91.93\% & 694 & - & 100\% \\
\midrule
\textbf{Ours (Gated Critic)} & \textbf{90.33\%} & \textbf{151} & 82 & 9.3\% \\
\bottomrule
\end{tabular}}
\end{table}
\subsection{Dataset Statistics of Condition and Condition Encoding Strategy}
\paragraph{Data Preprocessing and Featurization}
To explicitly model the influence of reaction conditions, we first disentangle the input precursors into core reactants and conditional agents based on atom-mapping information. Given a reaction SMILES string $S_{src} \gg S_{tgt}$, we parse the product $S_{tgt}$ to retrieve the set of mapped atom indices $\mathcal{I}_{tgt}$. For each molecule $m$ in the source $S_{src}$, we classify it as a reactant if any of its atoms possess a mapping index present in $\mathcal{I}_{tgt}$; otherwise, it is categorized as a conditional agent (e.g., solvent, catalyst, or reagent). The dataset construction proceeds in three steps:
\begin{itemize}\item \textbf{Role Separation:} The source molecules are split into a reactant graph $\mathcal{G}_R$ and a condition set $\mathcal{C}$.\item \textbf{Reactant Re-indexing:} Since removing condition molecules creates gaps in atom mapping indices, we perform a global re-indexing operation to map the remaining reactant atoms to a contiguous range $\{1, \dots, N_R\}$. This ensures consistency between the input graph and the output prediction targets.\item \textbf{Dual-view Featurization:}\begin{enumerate}\item \textbf{Graph Features:} Reactants are featurized into node (atom type, charge, aromaticity) and edge (bond type) matrices.\item \textbf{Condition Features:} The condition set $\mathcal{C}$ is processed into two formats: (i) a sequence of Molecular Fingerprints (ECFP4) for the permutation-invariant set encoder, and (ii) a global context vector combining a one-hot encoding of high-frequency agents (Top-128) 
\end{enumerate}
\end{itemize}

To justify our design choice of the semantic condition encoder ($h_{\text{set}}$), we analyzed the frequency distribution of all conditional agents (reagents, solvents, catalysts) in the training set.
As shown in  Table ~\ref{tab:condition_freq}, the data exhibits a significant long-tail distribution.
\begin{itemize}
\item \textbf{Head (High-frequency):} A small number of common solvents and ions dominate the dataset. For instance, $[Na^+]$, $H_2O$, and THF  appear in 21.1\%, 21.1\%, and 14.3\% of reactions, respectively.
\item \textbf{Tail (Low-frequency):} In contrast, the vast majority of agents are sparse. Highly specific catalysts and complex ligands (e.g., Palladium complexes, specialized phosphine ligands) often appear with a frequency of less than 0.01\% (see Table ~\ref{tab:condition_freq}).
\end{itemize}

\begin{table}[t]
\centering
\caption{\textbf{Frequency Statistics of Representative Conditional Agents.} The distribution is heavily skewed. Notably, a significant portion of agents are \textit{singletons} (appear only once), making standard learnable embeddings ineffective and necessitating chemically grounded fingerprint inputs.}
\label{tab:condition_freq}
\resizebox{\linewidth}{!}{%
\begin{tabular}{llcc}
\toprule
\textbf{Category} & \textbf{Representative SMILES} & \textbf{Count} & \textbf{Frequency (\%)} \\
\midrule
\multicolumn{4}{l}{\textit{Head (Common Solvents \& Ions)}} \\
$[Na^+]$ & $[Na^+]$ & 86,507 & 21.16 \\
$H_2O$ & O & 86,477 & 21.15 \\
THF & C1CCOC1 & 58,767 & 14.38 \\
\midrule
\multicolumn{4}{l}{\textit{Torso (Common Reagents)}} \\
Acetic Acid & CC(=O)O & 13,985 & 3.42 \\
Palladium & $[Pd]$ & 5,623 & 1.37 \\
$PdCl_2$ & Cl[Pd]Cl & 1,936 & 0.47 \\
\midrule
\multicolumn{4}{l}{\textit{Extreme Tail (Singletons) - Agents appearing exactly once}} \\
Bromo-methylpyridine & Cc1cccnc1Br & 1 & $\approx 0.0002$ \\
Pyrophosphate deriv. & O=P([O-])([O-])OP(=O)([O-])[O-] & 1 & $\approx 0.0002$ \\
Bicyclic Amine & NC1CC2CCN1CC2 & 1 & $\approx 0.0002$ \\
Complex Diazo & [N-]=[N+](C(=O)[O-])C(=O)[O-]  & 1 & $\approx 0.0002$ \\
\bottomrule
\end{tabular}
}
\end{table}
\end{document}